\documentclass[twoside]{article}

\usepackage{PRIMEarxiv}

\usepackage[utf8]{inputenc} 
\usepackage[T1]{fontenc}    
\usepackage{hyperref}       
\usepackage{url}            
\usepackage{booktabs}       
\usepackage{amsfonts}       
\usepackage{nicefrac}       
\usepackage{microtype}      
\usepackage{lipsum}
\usepackage{fancyhdr}       
\usepackage{graphicx}       
\graphicspath{{media/}}     
\usepackage{multirow}

\usepackage{newunicodechar}

\usepackage[most]{tcolorbox}
\usepackage{enumitem}
\usepackage[table]{xcolor}  
\usepackage[most]{tcolorbox}
\usepackage{xcolor}
\usepackage[normalem]{ulem}
\usepackage{pifont}
\newcommand{\cmark}{\ding{51}}
\newcommand{\xmark}{\ding{55}}
\newtcolorbox{promptbox}[1]{
    enhanced,
    breakable,
    width=\columnwidth,
    title=\textbf{#1},
    colback=gray!3,
    colframe=gray!40,
    colbacktitle=gray!10,
    coltitle=black,
    fonttitle=\small,
    fontupper=\footnotesize\ttfamily,
    boxrule=0.5pt,
    arc=2pt,
    left=5pt,
    right=5pt,
    top=5pt,
    bottom=5pt
}

\title{TRIDENT and SHAKTI: A Round-Trip Translation-Based Framework and Large-Scale Benchmark Dataset for AI-Generated Hindi Text Detection
}

\author{
  Shifali Agrahari \\
  Department of CSE, \\
   Indian Institute of Technology Guwahati \\
  Assam, India\\
  \texttt{a.shifali@iitg.ac.in} \\
   \And
    Sujit Kumar\\
  Institute of Advanced Intelligence  \\
  and Computing (IAIC),\\
   A*STAR\\
  Singapore\\
  \And
  Sanasam Ranbir Singh \\
   Department of CSE, \\
   Indian Institute of Technology Guwahati \\
  Assam, India\\
}

\begin{document}
\maketitle

\begin{abstract}
As Large Language Models (LLMs) continue to advance and become increasingly multilingual, distinguishing between human-written and LLM-generated text is crucial for preserving trust and integrity in information dissemination, public discourse, and digital ecosystems. However, existing research on AI-generated text detection remains predominantly English-centric, with limited availability of comprehensive benchmark datasets for low-resource languages such as Hindi, particularly those encompassing diverse text genres, LLM generators, prompting strategies, and adversarial transformations. To address this gap, we introduce SHAKTI, a large-scale Hindi benchmark for AI-generated text detection, featuring (i) human-written content across four genres : essays, news, Wikipedia, and question answering; (ii) AI-generated texts produced by eight LLM families using multiple prompting strategies; and (iii) adversarially transformed samples to evaluate detection robustness under both conventional and adversarial settings. Next, we assess the quality and reliability of SHAKTI through human annotation and evaluation, and systematically investigate differences between human-written and LLM-generated Hindi texts using linguistic, grammatical, and sentiment-based features. Building on these insights, we propose TRIDENT and its three variants, \textsc{TRIDENT-S} (Static Threshold-Based Classification), \textsc{TRIDENT-D} (Dynamic Threshold-Based Classification), and \textsc{TRIDENT-C} (Logistic Regression-Based Classification with Multi-Distance RTT Features), a round-trip translation-based detection framework comprising three complementary classification variants. TRIDENT exploits differences in semantic consistency between original and round-trip translated texts to distinguish human-written from LLM-generated content. Extensive experiments demonstrate that TRIDENT outperforms conventional statistical, rewriting-based, and supervised detection baselines across the evaluated Hindi datasets, with the static-threshold variant (\textsc{TRIDENT-S}) achieving the strongest overall performance. Furthermore, evaluations across Hindi, Urdu, Arabic, Russian, Chinese, English and Korean demonstrate its multilingual applicability while revealing sensitivity to the choice of translation model. Adversarial perturbation and cross-LLM generalization evaluations further assess its robustness and ability to detect text from unseen LLMs\footnote{The dataset is available upon reasonable request from the corresponding authors. The dataset and source code will be made publicly available upon publication of the peer-reviewed version of this work.}.
\end{abstract}


\section{Introduction}

The rapid development of Large Language Models (LLMs) has driven substantial progress across a range of natural language generation tasks, such as writing news articles~\cite{yanagi2020fake}, generating stories~\cite{yuan2022wordcraft}, and review generation~\cite{salminen2022creating}. As LLMs become increasingly accessible, the associated risks have also grown significantly. LLM-generated content can contribute to the proliferation of fake news and misinformation~\cite{wang2022fake}, scams by generating convincing malicious emails, and facilitation of academic dishonesty~\cite{mirsky2021creation, zellers2019defending, perkins2023academic}. Consequently, distinguishing human-written from LLM-generated text is essential for combating misinformation, preventing the dissemination of deceptive content, safeguarding academic integrity, and preserving trust in digital information ecosystems.
\begin{figure}[t]
\centering 
\includegraphics[width=\textwidth]{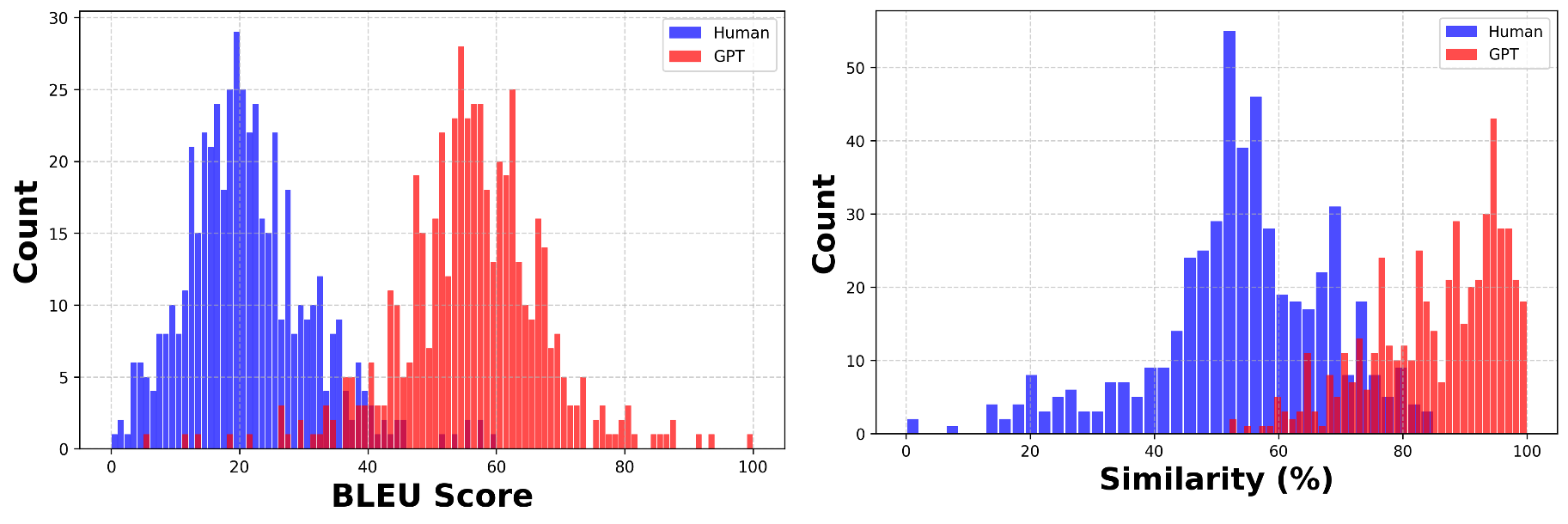}  
\caption{
\textbf{RTT (Round Trip Translations) consistency for \textcolor{blue}{Human-written} and \textcolor{red}{GPT-generated text}.} The \textbf{left panel} shows the distribution of BLEU scores, with the \textbf{x-axis} representing BLEU scores (0--100) and the \textbf{y-axis} representing the number of samples in each bin. The \textbf{right panel} shows the distribution of cosine similarity scores between the original and round-trip-translated texts (Hindi $\rightarrow$ English $\rightarrow$ Hindi), with the \textbf{x-axis} representing cosine similarity expressed as a percentage (0--100\%) and the \textbf{y-axis} representing the number of samples in each bin. GPT-generated text generally exhibits higher lexical and semantic consistency after round-trip translation, whereas human-written text shows greater translation-induced variation. These distributional differences suggest that RTT consistency may provide a useful signal for distinguishing AI-generated from human-written text.}
\label{fig:roundtrip_consistency} 
\end{figure}

Despite its importance, research on the detection of LLM-generated text in Hindi remains extremely limited, and according to Wikipedia (2023)\cite{wikipediaHindi}, Hindi is the fourth most spoken language in the world\footnote{\url{https://www.ethnologue.com/insights/most-spoken-language/}}. To date, relatively few studies have investigated LLM-generated text detection involving Hindi. The $AG_{\text{hi}}$ dataset~\cite{kavathekar2024counter} focuses on Hindi news articles, whereas IndicDetect~\cite{devalla2026indicdetect} extends the investigation to Hindi, Telugu, and Tamil across multiple domains, including news, journal articles, poetry, and entertainment content. Despite these contributions, existing datasets differ in scale, domain coverage, generator diversity, and support for adversarial evaluation. In particular, the availability of large-scale Hindi benchmarks incorporating diverse LLM generators, including models specifically developed for Indian languages, remains limited. Furthermore, although existing datasets incorporate various text perturbations, there remains a need for comprehensive evaluation of detection robustness across diverse adversarial transformations and generation settings. Furthermore, real-world scenarios often involve adversarial attacks \cite{kadhim2025adversarial}, where malicious actors intentionally manipulate text such as rewrite, synonym substitution, or grammatical changes to evade LLM-generated text detectors. These attacks undermine detection robustness, necessitating their consideration in the dataset and evaluation of LLM generated text detector. 

To address this gap, we introduce \textit{SHAKTI} (Source-aware Hindi AI-Generated Text Identification), a large-scale Hindi benchmark comprising human-written, LLM-generated, and adversarially perturbed texts across four genres: news, question answering, Wikipedia, and essays. SHAKTI incorporates outputs from diverse LLMs, including GPT-3.5~\cite{radford2019language}, Meta-LLaMA-3.2-70B~\cite{touvron2023llama}, Mistral-Nemo-Base-2407~\cite{jiang2023mistral}, Gemma~\cite{team2024gemma}, ChatGPT (GPT-5)\footnote{\url{https://chatgpt.com/?model=auto}} and Gemini\footnote{\url{https://gemini.google.com/app}}, and Indian-language-focused models such as Aya 101 and Sarvam-1 AI\footnote{\url{https://huggingface.co/sarvamai/sarvam-1}}. By covering diverse generation settings and adversarial transformations, SHAKTI enables systematic evaluation of AI-generated text detection in Hindi.

Despite significant advances in LLM-generated text detection~\cite{fagni2021tweepfake, gehrmann2019gltr, ippolito2019automatic, yu2024fake}, reliably distinguishing human-written from LLM-generated text across diverse domains and distributions remains an open challenge.  In particular, supervised detectors~\cite{solaiman2019release, fagni2021tweepfake, su2023detectllm, liyanage2024detecting, liu2023argugpt} often achieve strong in-domain performance but experience substantial degradation under out-of-distribution (OOD) conditions, highlighting their limited generalizability across unseen domains and text distributions. Zero-shot methods~\cite{bao2024fast, gehrmann2019gltr, mitchell2023detectgpt, su2023detectllm, ma2024zero} often rely on LLM logit outputs, which are unavailable for most black-box commercial models such as GPT-3.5 and GPT-4. Additionally, high-dimensional feature-based detectors may include redundant or spurious features, leading to overfitting. Recent approaches, including SimLLM~\cite{nguyen2024simllm}, RAIDAR~\cite{mao2024raidar}, and LLaMA-L2R~\cite{hao-etal-2025-learning}, detect AI-generated text by rewriting the input and comparing the rewritten version with the original. However, rewriting is an unconstrained transformation that alters meaning and depends on the rewriting model, making similarity unstable.

Motivated by these limitations, we propose \textbf{TRIDENT}  (\textbf{T}ranslation-based \textbf{R}ound-trip \textbf{I}dentification for \textbf{D}etecting \textbf{EN}tities of \textbf{T}ext), a novel framework that leverages translation-induced variations through round-trip translation (RTT) to distinguish human-written from LLM-generated text. The primary instinct behind our proposed TRIDENT methods is that LLM-generated text, when translated and back-translated via LLM only, naturally reinforces its own stylistic and structural tendencies, resulting in high similarity between the original LLM-generated text and its RTT version. A plausible explanation is that, during translation and back-translation, the LLM tends to reconstruct text according to its learned generative patterns, potentially preserving stylistic characteristics associated with LLM-generated content. In contrast, human-written text demonstrates greater stylistic richness and less predictable structure.  During a round-trip translation using an LLM, the LLM tends to reshape the text toward its preferred writing style, reducing the similarity between the original human-written text and its corresponding RTT output. This difference in similarity patterns enables reliable discrimination between human-generated and LLM-generated text. TRIDENT preserves semantic meaning while perturbing surface form, producing a more robust signal for LLM-generated text detection. 

Figure~\ref{fig:roundtrip_consistency} illustrates the distribution of BLEU scores and similarity between the original text and its RTT. As shown in Figure~\ref{fig:roundtrip_consistency}, GPT-generated text exhibits substantially higher BLEU scores and cosine similarity following RTT compared to human-written text. The pronounced separation in BLEU distributions suggests that GPT-generated text undergoes fewer lexical changes, while the higher cosine similarity indicates stronger semantic preservation. Although some overlap exists, particularly in semantic similarity, these observations suggest that human-written and LLM-generated texts exhibit distinct responses to translation-induced transformations. Motivated by the above observation from  Figure~\ref{fig:roundtrip_consistency}, we exploit RTT consistency as a discriminative signal for detecting LLM-generated text in our proposed TRIDENT framework.

We develop three complementary variants of TRIDENT that differ in their classification strategies. TRIDENT-S (Static Threshold) uses a fixed decision threshold selected using labelled validation data. TRIDENT-D (Dynamic Threshold) estimates an adaptive decision boundary using an unsupervised two-component Gaussian Mixture Model (GMM) fitted to unlabelled validation scores. {TRIDENT-C (Logistic Regression) employs a supervised logistic regression classifier trained on multiple RTT-derived features, capturing complementary signals of lexical, semantic, and distributional variation. Together, these variants enable the investigation of fixed-threshold, adaptive-threshold, and supervised approaches to RTT-based detection.

We conduct extensive experiments on our proposed \textsc{SHAKTI} benchmark and publicly available datasets from literature, spanning eight languages. We compare the performance of \textsc{TRIDENT} against representative state-of-the-art methods  from the literature, both in-domain and cross-domain evaluation settings. The key findings from our experiments demonstrate the effectiveness and generalizability of \textsc{TRIDENT} for detecting LLM-generated text. \textsc{TRIDENT} consistently outperforms conventional statistical, rewriting-based, and supervised detection baselines across the evaluated Hindi datasets and genres. Furthermore, \textsc{TRIDENT} demonstrates promising multilingual applicability, achieving strong detection performance across datasets in seven languages other than Hindi, although its effectiveness varies across languages and translation models.  Our findings also reveal that the choice of translation model significantly influences detection performance, with LLM-based translators generally outperforming conventional machine translation models and GPT-3.5 exhibiting strong performance across multiple settings. Our key contributions can be summarized as follows:
\begin{enumerate}
    \item We curate \textbf{SHAKTI}, a large-scale Hindi benchmark to foster the development and evaluation of methods for LLM-generated text detection in Hindi, comprising human-written, LLM-generated, and adversarially transformed texts across diverse domains and genres, and from multiple LLM generators. Our proposed \textbf{SHAKTI} dataset facilitates systematic evaluation of detection methods under diverse generation and perturbation settings.
    \item We propose \textbf{TRIDENT}, a round-trip translation (RTT)-based framework for detecting LLM-generated text across multiple languages. TRIDENT comprises three complementary variants: TRIDENT-S, which employs static threshold-based classification; TRIDENT-D, which uses dynamic threshold estimation based on an unsupervised Gaussian Mixture Model (GMM); and TRIDENT-C, which employs supervised logistic regression on RTT-derived features.
    \item We conduct extensive experiments on \textsc{SHAKTI} and existing benchmark datasets across multiple languages, evaluating the effectiveness and multilingual applicability of \textsc{TRIDENT} against representative statistical, rewriting-based, and supervised detection methods.
    \item We provide a comprehensive empirical analysis of RTT-induced consistency, examining the influence of translation models, text genres, and classification strategies on detection performance. We further investigate linguistic characteristics and misclassification patterns to identify the strengths and limitations of RTT-based detection.

\end{enumerate}

\begin{table*}[t]
\centering
\caption{Comparison of existing AI-generated text detection benchmarks
in terms of dataset size, language coverage, domain diversity, generator
diversity, adversarial robustness, and language-specific LLM coverage.
Language-specific LLMs refer to models developed or adapted for regional
languages, such as Sarvam models for Indian languages. General-purpose
models include GPT, Llama, and Qwen, while Aya is treated as a multilingual
model family. SHAKTI provides a large-scale Hindi benchmark with multiple
domains, diverse generators, and five adversarial transformations.}
\label{tab:benchmark_comparison}

\renewcommand{\arraystretch}{1.15}
\setlength{\tabcolsep}{4pt}
\rowcolors{2}{gray!30}{white}

\resizebox{\textwidth}{!}{%
\begin{tabular}{l l c c c c c}
\toprule
\rowcolor{blue!50!black}
\textcolor{white}{\textbf{Name}} &
\textcolor{white}{\textbf{Size}} &
\textcolor{white}{\textbf{Langs}} &
\textcolor{white}{\textbf{Multi-Domain}} &
\textcolor{white}{\textbf{Multi-Generator}} &
\textcolor{white}{\textbf{Adversarial}} &
\textcolor{white}{\textbf{Language-Specific LLMs}} \\
\midrule

TuringBench~\cite{uchendu2021turingbench}
& 200k & EN & \xmark & \cmark & \xmark & \xmark \\

RuATD~\cite{shamardina2022findings}
& 215k & RU & \cmark & \cmark & \xmark & \xmark \\

HC3~\cite{guo2023close}
& 26.9k & EN/ZH & \cmark & \xmark & \xmark & \xmark \\

MGTBench~\cite{he2024mgtbench}
& 2.8k & EN & \cmark & \cmark & \cmark & \xmark \\

MULTITuDE~\cite{macko2023multitude}
& 74.1k & Multi & \cmark & \cmark & \xmark & \xmark \\

M4~\cite{wang2023m4}
& 122k & Multi & \cmark & \cmark & \xmark & \xmark \\

CCD~\cite{wang2023evaluating}
& 467k & EN & \xmark & \cmark & \cmark & \xmark \\

IMDGSP~\cite{mosca2023distinguishing}
& 29k & EN & \xmark & \cmark & \xmark & \xmark \\

HC-Var~\cite{xu2024generalization}
& 145k & EN & \cmark & \xmark & \xmark & \xmark \\

HC3 Plus~\cite{su2023hc3}
& 210k & EN/ZH & \cmark & \xmark & \xmark & \xmark \\

MAGE~\cite{li2024mage}
& 447k & EN & \cmark & \cmark & \xmark & \xmark \\

HC3-French~\cite{antoun2023towards}
& 54k & FR & \xmark & \xmark & \cmark & \xmark \\

DetectRL~\cite{wu2024detectrl}
& 1.4M & EN & \cmark & \cmark & \cmark & \xmark \\

RAID~\cite{dugan2024raid}
& 6.2M & EN & \cmark & \cmark & \cmark & \xmark \\

BUST~\cite{cornelius2024bust}
& 25k & EN & \cmark & \cmark & \xmark & \xmark \\

DetectRL-X~\cite{wu2026detectrl}
& 3.46M & Multi & \cmark & \cmark & \cmark & \xmark \\

MultiSocial~\cite{macko2024multisocial}
& 414k & Multi & \cmark & \cmark & \xmark & \xmark \\

AGhi~\cite{kavathekar2024counter}
& 36.7k & HI & \xmark & \cmark & \xmark & \xmark \\

IndicDetect~\cite{devalla2026indicdetect}
& 84k & HI/TE/TA & \cmark & \cmark & \cmark & \xmark \\

HLU~\cite{ali2025hlu}
& 1,014 & UR & \cmark & \xmark & \xmark & \xmark \\

KatFishNet~\cite{park2025katfishnet}
& 2,094 & Korean & \cmark & \cmark & \xmark & \xmark \\

Arabic~\cite{al2025arabic}
& 3,500 & Arabic & \cmark & \cmark & \xmark & \xmark \\

\midrule
\rowcolor{gray!50}
\textbf{SHAKTI (Ours)}
& \textbf{1.01M}
& \textbf{HI}
& \textbf{\cmark}
& \textbf{\cmark}
& \textbf{\cmark}
& \textbf{\cmark} \\

\bottomrule
\end{tabular}%
}

\vspace{3pt}
\begin{minipage}{0.98\textwidth}
\footnotesize
\textit{Notes:}
EN = English; ZH = Chinese; RU = Russian; FR = French;
HI = Hindi; TE = Telugu; TA = Tamil; UR = Urdu;
Multi = multiple languages.
A checkmark indicates that the capability is supported in this
comparison, while a cross indicates that it is not indicated as
supported. SHAKTI contains 1,006,800 samples across human-written,
LLM-generated, and adversarial categories. Its five adversarial
transformations are rewriting, homoglyph substitution, synonym
replacement, misspelling, and whitespace manipulation.
Benchmark statistics and capability indicators should be verified
against the original publications before submission.
\end{minipage}
\end{table*}

\section{Related Work}
In literature studies~\cite{ahmad2026human, sharimbayev2026ai, attar2026systematic, wu2025survey,fraser2025detecting,mirsky2021creation} have briefly reviewed works related to AI-generated content in the English language. In contrast, this section provides a focused review of existing datasets and methodological approaches for detecting AI-generated text in Hindi, highlighting the progress and research gaps in this relatively underexplored language.

\subsection{AI-Generated Text Dataset} 

The rapid advancement of large language models (LLMs) has motivated the development of datasets and benchmarks for distinguishing human-written text from machine-generated content. Existing benchmarks differ in their language coverage, domain diversity, generator composition, and evaluation protocols, making direct comparisons across detection methods challenging~\cite{wu2025survey,sharimbayev2026ai,ahmad2026human}. Although several datasets have enabled systematic evaluation of AI-generated text detectors, limitations in linguistic diversity, distributional coverage, and robustness evaluation remain important challenges. One of the early contributions to AI-generated text detection, the TuringBench dataset~\cite{uchendu2021turingbench}, provides large collections of human-written and machine-generated news articles produced by multiple language models. The MAGE dataset~\cite{li2024mage} extends this line of research by incorporating diverse text sources and generation settings to investigate the generalisation capabilities of AI-generated text detectors. However, both benchmarks primarily focus on English, limiting their ability to characterise detection performance across languages with different linguistic structures and writing conventions~\cite{wu2025survey,macko2023multitude}.

Subsequent studies, RAID~\cite{dugan2024raid} and DetectRL~\cite{wu2024detectrl} mainly emphasise evaluating detector robustness under realistic distribution shifts. RAID~\cite{dugan2024raid} provides a large-scale benchmark dataset incorporating samples generated by multiple generators (LLMs) over a corpus from diverse domains and   decoding strategies, with different perturbation techniques, including paraphrasing and other text transformations. RAID~\cite{dugan2024raid} provides a large-scale benchmark dataset incorporating samples generated by multiple generators (LLMs) over a corpus from diverse domains and   decoding strategies, with different perturbation techniques, including paraphrasing and other text transformations. While DetectRL~\cite{wu2024detectrl} further investigates detection under challenging generation and text-transformation settings. These resources facilitate training and evaluation of models beyond conventional in-distribution evaluation for AI-generated text detection. Nevertheless, their predominantly English-language focus limits their applicability to multilingual settings, where linguistic variation and language-specific transformations may introduce additional challenges~\cite{wu2025survey,macko2023multitude}.

To address the limitations of predominantly English-centric datasets, subsequent studies have introduced multilingual benchmarks that extend AI-generated text detection research to a broader range of languages. The M4 dataset~\cite{wang2023m4} extends machine-generated text detection to multilingual, multi-domain, and multi-generator settings, enabling the evaluation of detector robustness and generalisation across heterogeneous data distributions. However M4 dataset~\cite{wang2023m4}, imbalances in sample sizes, domain distributions, and generator coverage across languages introduce distributional confounders, limiting the reliability of cross-lingual performance comparisons. HC3~\cite{guo2023close} introduces a human–ChatGPT comparison corpus spanning English and Chinese across multiple question-answering domains, facilitating the evaluation of human-versus-machine text detection. Extending multilingual evaluation, MULTITuDE~\cite{macko2023multitude} comprises human-written and machine-generated news texts across eleven languages and eight LLMs, enabling systematic investigation of cross-lingual and cross-generator generalisation. However, these benchmarks provide limited systematic evaluation of adversarial text transformations and cross-domain distribution shifts, leaving detector robustness under realistic out-of-distribution conditions insufficiently explored.

Furthermore, SemEval-2024 Task 8~\cite{wang2024semeval} established evaluation tracks for machine-generated text detection and related attribution tasks, including multilingual settings. The multilingual GenAI content detection task~\cite{wang2025genai} further advances the evaluation of machine-generated content across languages. However, the availability of dedicated, sufficiently diverse datasets remains important for evaluating robustness beyond the specific languages, domains, and generators represented in shared-task test sets. Beyond multilingual benchmarks, language-specific studies have investigated AI-generated text detection across diverse linguistic settings. HLU~\cite{ali2025hlu} introduces multi-granularity detection benchmarks for Urdu, while KatFishNet~\cite{park2025katfishnet} exploits language-specific morphological and syntactic features for Korean text detection. Similarly, AuTexTification~\cite{sarvazyan2023overview} investigates machine-generated text detection and generator attribution across multiple domains in Spanish and English, whereas RuATD~\cite{shamardina2022findings} establishes a benchmark for AI-generated text detection in Russian. Studies~\cite{ali2025hlu,park2025katfishnet, sarvazyan2023overview, shamardina2022findings, kavathekar2024counter} underscore the importance of language-specific evaluation, particularly in capturing morphological, syntactic, and sociolinguistic variations that influence AI-generated text detection. However, heterogeneity in dataset size, domain distributions, generator diversity, and adversarial perturbation coverage introduces evaluation biases, limiting systematic assessment of detector robustness and cross-lingual generalization. 

In the context of Hindi, the Counter Turing Test (CT\textsuperscript2) study~\cite{kavathekar2024counter} introduces AG$_{\text{hi}}$, a language-specific benchmark for training and evaluation of models for AI-generated text detection in Hindi language. The AG$_{\text{hi}}$~\cite{kavathekar2024counter} dataset comprises human-written news articles sourced from BBC Hindi and NDTV, alongside corresponding AI-generated news  articles generated by five LLMs. Nevertheless, AG$_{\text{hi}}$ is limited to news articles sourced from BBC Hindi and NDTV, alongside machine-generated texts produced by five LLMs using headline-conditioned generation. This restricts its diversity in terms of text domains, writing styles, source distributions, and generation strategies. Furthermore, imbalances in generator-specific samples, limited variation in decoding configurations, and the absence of systematic adversarial transformations, such as paraphrasing, human post-editing, and mixed human–AI authorship, limit its ability to comprehensively evaluate detector robustness and generalization across diverse text conditions and real-world distribution shifts.

To address these limitations, we introduce SHAKTI, a large-scale Hindi AI-generated text detection benchmark covering four domains: essays, news, Wikipedia, and question answering. It includes human-written and AI-generated texts from diverse LLMs, including GPT-3.5, LLaMA-3.2, Mistral, Gemma-3, GPT-5, Gemini-2.5 Flash, Aya, and Sarvam. The dataset incorporates adversarial transformations, including rewriting, homoglyph substitution, synonym replacement, misspelling, and whitespace manipulation, alongside diverse text lengths, prompts, and prompting strategies. SHAKTI enables systematic evaluation of cross-domain and cross-generator generalization and detector robustness against text modifications. Table~\ref{tab:benchmark_comparison} compares SHAKTI with existing datasets in terms of scale, domain and generator diversity, LLM operations, language coverage, and adversarial evaluation.

\subsection{Detection Methods} 
Existing AI-generated text detection methods can be broadly classified into three major categories: (i) statistical and zero-shot detection methods, (ii) supervised learning-based detection methods, and (iii) rewriting-based detection methods.

\textbf{Statistical and zero-shot methods :}
These approaches exploit statistical properties of text to detect AI-generated content, often without requiring task-specific labelled training data. White-box-based methods leverage language-model probability distributions to estimate statistical signals, including perplexity~\cite{beresneva2016computer}, likelihood~\cite{solaiman2019release}, and token ranks~\cite{gehrmann2019gltr}. DetectGPT~\cite{mitchell2023detectgpt} exploits changes in log-probability under text perturbations, while Fast-DetectGPT~\cite{bao2024fast} improves computational efficiency through conditional probability curvature estimation. DetectLLM~\cite{su2023detectllm} employs log-rank-based statistics, whereas Binoculars~\cite{hans2401spotting} leverages the relationship between perplexity and cross-perplexity computed using two language models. The DNA-GPT~\cite{yanagi2020fake} study examines consistency between human-authored text and AI-generated text, while Biscope~\cite{guo2024biscope} investigates the role of preceding-token information in AI-generated content detection.

\textbf{Supervised learning-based methods:} These methods use labeled human-written and AI-generated texts to learn discriminative linguistic, stylistic, and semantic representations. Early approaches fine-tune pretrained language models for binary classification~\cite{solaiman2019release}, while Rivera et al.~\cite{rivera2024few} leverage human writing-style representations to improve cross-generator generalization. RADAR~\cite{hu2023radar} employs adversarial detector–paraphraser training, and OUTFOX~\cite{koike2024outfox} uses adversarial examples with iterative in-context learning. Ghostbuster~\cite{verma2024ghostbuster} combines probability estimates from weaker language models to construct discriminative features, whereas EAGLE~\cite{bhattacharjee2024eagle} uses contrastive learning and domain-adversarial training to improve generalization to unseen generators. DeTeCtive~\cite{guo2024detective} integrates multi-level contrastive learning and dense retrieval to capture author-specific writing styles. Beyond conventional in-domain evaluation, Wang et al.~\cite{wang2025genai} introduced a shared-task benchmark for multilingual machine-generated text detection. Liyanage et al.~\cite{liyanage2024detecting} investigated LLM-generated hotel review detection using traditional machine-learning and fine-tuned transformer models, highlighting the effects of training-data availability and paraphrasing. ArguGPT~\cite{liu2023argugpt} provides a benchmark of human-written and GPT-generated argumentative essays, demonstrating effective detection using fine-tuned RoBERTa at essay and sentence levels. In educational applications, Xiao et al.~\cite{xiao2025human} proposed a dual-process-inspired LLM framework for automated essay scoring that integrates AI-generated evaluations, explanatory feedback, and confidence estimates to support human–AI collaborative grading.

\textbf{Rewriting-based methods :} These approaches exploit differences in the rewriting behavior of human-written and AI-generated texts, leveraging changes in linguistic structure, semantic consistency, or model likelihood to distinguish their origins. SimLLM~\cite{nguyen2024simllm} exploits similarity between input sentences and their regenerated counterparts, while DART~\cite{park2024dart} leverages Abstract Meaning Representation (AMR) of rephrased texts to capture semantic differences for detection. Building on the observation that LLMs tend to modify human-written text more extensively than machine-generated content, RAIDAR~\cite{mao2024raidar} employs edit distance between original and rewritten texts as a detection signal. Extending this principle, LLaMA-L2R~\cite{hao-etal-2025-learning} trains a rewriting model to amplify these differences, improving generalization across domains and generation settings. Nevertheless, the effectiveness of rewriting-based approaches may depend on the rewriting model, transformation strategy, and characteristics of the input text. Despite promising results, existing AI-generated text detection methods face several challenges. First, generalization to unseen generators and domains remains difficult: evaluations on the M4 benchmark show that detectors can misclassify machine-generated text from previously unseen LLMs and domains~\cite{wang2023m4}. Second, rewriting-based methods depend on the behaviour of the rewriting model and the transformation applied. Methods such as RAIDAR use the edit distance between original and rewritten texts, while SimLLM compares texts with their regenerated counterparts~\cite{mao2024raidar, nguyen2024simllm}. Consequently, the discriminative value of these signals may vary with the rewriting or regeneration model and the characteristics of the input text. Third, some approaches require additional training or model-specific resources. For example, Learning2Rewrite trains a rewriting model to amplify the differences between human-written and LLM-generated text, introducing additional training requirements~\cite{hao-etal-2025-learning}. Fourth, robustness across languages, domains, and evolving generation strategies remains challenging. Although rewriting-based methods have demonstrated promising generalization, their effectiveness should be assessed under shifts in language, domain, and generator rather than assumed to transfer universally~\cite{wang2023m4, macko2023multitude}. These limitations motivate the development of detection approaches that exploit complementary signals and reduce dependence on task-specific detector training. Despite their strong performance in controlled evaluation settings, supervised detection methods often exhibit limited generalization to out-of-distribution scenarios, particularly when encountering unseen language models, domains, and languages~\cite{wang2023m4} Their effectiveness largely depends on the diversity and representativeness of labeled training data, making them vulnerable to distribution shifts arising from evolving language models and text-generation strategies. Furthermore, although statistical and zero-shot approaches reduce reliance on task-specific labeled data, their effectiveness can be constrained by access to language-model probability distributions and sensitivity to text characteristics and transformations. Consequently, developing robust detection methods that generalize across generators, domains, and linguistic settings remains an important research challenge. Rewriting-based methods face three key limitations: (i) poor out-of-domain generalisation, (ii) dependence on labelled training data, and (iii) sensitivity to evolving LLM behaviours. 

Motivated by these challenges, we propose \textbf{TRIDENT}, a round-trip translation-based detection framework with three variants that exploit translation-induced textual changes to distinguish human-written from LLM-generated content. We also introduce \textbf{SHAKTI}, a Hindi benchmark encompassing multiple domains, diverse LLM generators, and adversarially transformed texts. Together, TRIDENT and SHAKTI provide a complementary framework for developing and systematically evaluating AI-generated text detectors under diverse linguistic, distributional, and adversarial conditions.

\begin{figure}[t]
\small
    \centering
    \includegraphics[width=\linewidth]{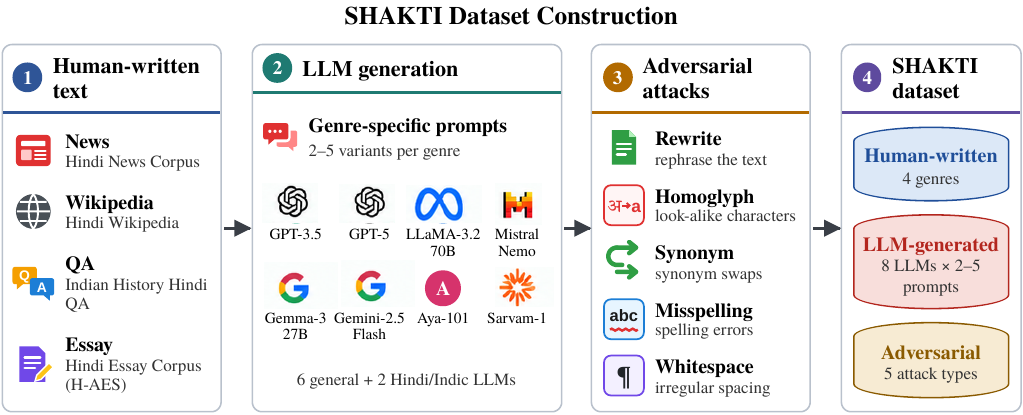} 
    \caption{Overview of the SHAKTI dataset construction pipeline. Human-written Hindi texts are collected from four genres—news, Wikipedia, question answering (QA), and essays—and used to guide text generation with eight LLMs and multiple genre-specific prompts. Five adversarial transformations are subsequently applied to LLM-generated texts to construct a benchmark comprising human-written, unperturbed LLM-generated, and adversarially transformed samples.}
    \label{fig:first_image}
\end{figure}

\section{Proposed Dataset}
\subsection{Human-Written Text Collection}
We construct SHAKTI using human-written Hindi texts collected from publicly available sources across four genres: (i) news~\cite{kumar2024fake}, (ii) Wikipedia~\cite{wikipediaHindiHome}, (iii) question answering (QA)~\cite{kaifahmad2024discussions}, and (iv) essays~\cite{singh2023h}. These genres represent different writing styles, text lengths, and levels of linguistic complexity, enabling a more comprehensive evaluation of AI-generated text detection. Appendix Figure~\ref{fig:dataset}  illustrates sample instances from the proposed dataset across four genres. The dataset exhibits substantial variation in sentence length, style, and textual complexity.

\begin{itemize}
\item \textbf{News:} We use the raw Hindi news corpus from a curated fake news detection dataset~\cite{kumar2024fake}, without using its fake news labels. The corpus comprises human-written articles collected from multiple media outlets and covers topics such as politics, sports, finance, current affairs, and daily news. Study~\cite{kumar2024fake} considers news articles collected from mainstream media outlets as True news articles and generated fake news articles using data augmentation techniques. In our study, we use True news articles from the study corpus~\cite {kumar2024fake} as human-authored news samples.
\item \textbf{Wikipedia:} We collected human-written Hindi encyclopedic articles from the Hindi Wikipedia dump (\texttt{hiwiki}) provided by Wikimedia Dumps\footnote{\url{https://dumps.wikimedia.org/hiwiki/}}. These articles cover diverse topics and provide informative, knowledge-oriented text, complementing other genres in our dataset.

\item \textbf{Question Answering (QA):} This genre comprises concise, explanatory, and conversational question--answer pairs. We sourced the Hindi (\texttt{hi}) subset from the AI4Bharat IndicQA dataset\footnote{\href{https://huggingface.co/datasets/ai4bharat/IndicQA}{AI4Bharat IndicQA dataset}}, which contains four fields: \textit{ID}, \textit{Context}, \textit{Question}, and \textit{Answers}. We exclusively extracted the question and answer fields to construct the QA genre.

\item \textbf{Essay:} We utilize the naturally collected Hindi essay
corpus introduced in H-AES~\cite{singh2023h}.
The corpus comprises 126 essays authored by students aged
18--20 as part of an online essay-writing competition. The
essays address a common travel-related topic, providing
authentic Hindi writing samples that reflect human linguistic
choices and natural variations in writing style. We use the
original Hindi essays and exclude the ASAP-Hindi subset, which
comprises English essays translated into Hindi.
\end{itemize}
The inclusion of news articles, Wikipedia content, question-answer texts, and essays introduces diversity in vocabulary, syntactic structures, discourse organisation, and text length.
\subsection{LLM-Generated Text Curation}
To curate the LLMs generated portion of SHAKTI, we use multiple LLMs representing different model families and generation capabilities.We consider six open-source models which inclide GPT-3.5~\cite{radford2019language}, Meta-LLaMA-3.2-70B~\cite{touvron2023llama}, Mistral-Nemo-Base-2407~\cite{jiang2023mistral}, and Gemma-3-27B-IT~\cite{team2024gemma}, ChatGPT~(GPT-5) and Gemini-2.5 Flash and two Hindi LLMs including Aya-101~\cite{ustun2024aya} and Sarvam-1\footnote{\url{https://huggingface.co/sarvamai/sarvam-1}}. We provide genre-specific instructions to generate Hindi text corresponding to the four genres described above. For each model-genre combination, we employ prompt variants to introduce diversity in tone, format, and response complexity. These variants include formal and conversational instructions, short and detailed responses, and different levels of explanatory depth. The prompts are designed to elicit varied outputs while maintaining consistency with the intended genre.

\subsection{Adversarial Text Generation}
In addition to unperturbed LLM-generated text, \textsc{SHAKTI} incorporates five types of adversarial transformations: rewriting~\cite{krishna2023paraphrasing}, homoglyph substitution~\cite{gagiano2021robustness}, synonym replacement~\cite{pu2023deepfake}, misspelling, and whitespace manipulation.These transformations are motivated by the need to evaluate whether detectors remain effective when machine-generated text is modified to obscure its original characteristics. The broader use of text perturbations in detection benchmarks is also explored in RAID~\cite{dugan2024raid}. The transformations are defined and applied as follows: 
\begin{itemize}
    \item \textbf{Rewrite Attack:} Rewrites the original text using different sentence structures or expressions while preserving its intended semantic meaning.

    \item \textbf{Homoglyph Attack:} Replaces selected characters with visually similar characters from the same or different character sets, introducing character-level perturbations while maintaining visual similarity.

    \item \textbf{Synonym Replacement Attack:} Replaces selected words with their synonyms to alter the lexical representation of the text while attempting to preserve its original meaning.

    \item \textbf{Misspelling Attack:} Introduces spelling errors, character insertions, deletions, substitutions, or transpositions to perturb the original text.

    \item \textbf{Whitespace Addition Attack:} Inserts additional or irregular whitespace within or between words to modify the textual representation without necessarily changing its intended meaning.
\end{itemize}
Appendix Figure~\ref{fig:adversial_example} presents five types of adversarial text
transformations in Hindi, including rewrite, homoglyph, and synonym replacement,
misspelling, and whitespace addition attacks.
The transformations are generated using language-model-based prompts or scripted perturbation operations, depending on the attack type. For rewriting, the original passage is provided to a language model instructed to rephrase it while preserving its central meaning. For synonym replacement, selected words are replaced with contextually appropriate alternatives, leaving the surrounding text unchanged wherever possible. Homoglyph substitution replaces eligible characters using a predefined mapping of visually similar characters. Misspelling generation applies the character-edit operations supported by the dataset construction pipeline to selected words. Whitespace manipulation inserts additional spaces at selected positions between or, where applicable, within words, while preserving the order of non-whitespace characters. The resulting perturbed samples are used to evaluate detector robustness against linguistic, character-level, and formatting modifications.

\subsection{Prompt Diversity and Reproducibility}
We use genre-specific prompt templates to generate Hindi text corresponding to four genres: news, Wikipedia-style encyclopedic text, question answering (QA), and essays. For each model--genre combination, we employ variants to introduce diversity in tone, format, and response complexity while maintaining consistency with the target genre. The corresponding templates specify the task, input, generation instructions, and expected output format.
To evaluate the robustness of AI-generated text detection, we further incorporate five adversarial transformations: rewrite, homoglyph substitution, synonym replacement, misspelling, and whitespace manipulation. We provide the corresponding attack prompts and transformation instructions to clarify how each perturbation is applied. While rewrite and synonym replacement modify lexical or syntactic expression, misspelling and whitespace manipulation introduce surface-level changes, and homoglyph substitution replaces selected characters with visually similar alternatives. These transformations are intended to preserve the original semantic content as much as possible while introducing different forms of textual variation.
The prompt templates and adversarial transformation instructions are presented in the following subsections. Together, they document the text-generation and perturbation procedures used to construct \textsc{SHAKTI} and facilitate reproducibility. Appendix Section~\ref{Appendix:propmt} provides representative examples of the prompt instructions used for curating the SHAKTI dataset. In summary, \textsc{SHAKTI} provides a comprehensive and diverse Hindi benchmark that integrates human-authored and LLM-generated texts across multiple genres, incorporates text authored by  diverse families of LLMs, and systematically includes five widely used adversarial transformation strategies. With its coverage of diversity in linguistic characteristics, writing styles, text lengths, generation sources, and adversarial perturbations, \textsc{SHAKTI} enables rigorous, systematic training and evaluation of AI-generated text detection methods across diverse textual conditions, including both conventional and adversarial settings.

\begin{table*}[t]
\centering
\caption{Comparison of Hindi AI-generated text detection datasets
in terms of domain coverage, data scale, LLM diversity, and
adversarial robustness.}
\label{tab:dataset_comparison}

\renewcommand{\arraystretch}{1.2}
\setlength{\tabcolsep}{5pt}
\small
\rowcolors{2}{gray!25}{white}

\resizebox{\textwidth}{!}{%
\begin{tabular}{p{3.0cm}p{4.0cm}p{3.6cm}p{4.4cm}}
\toprule
\rowcolor{blue!50!black}
\textcolor{white}{\textbf{Dimension}} &
\textcolor{white}{\textbf{$AG_{\text{hi}}$~\cite{kavathekar2024counter}}} &
\textcolor{white}{\textbf{IndicDetect~\cite{devalla2026indicdetect}}} &
\textcolor{white}{\textbf{SHAKTI (Ours)}} \\
\midrule

Language coverage
& Hindi (BBC/ NDTV)
& Hindi/Telugu/Tamil
& Hindi \\

Domain coverage
& News
& Google News, IJHR Journal, Poetry, Bollywood Hungama 
& Essay, news, Wikipedia, QA \\

LLM diversity
& 26 LLM, GPT versions, Bloom, BARD, mGPT, Mistral, Gemma, mT5 versions
& GPT-4.1, Qwen-Plus, DeepSeek-v3.2
& GPT3.5, Llama 3.2, Mistral, Gemma 3, GPT-5, Gemini, \textbf{Aya}, \textbf{Sarvam}\\

Attack coverage
& ---
& Perturbation, Whitespace, Insert-Paragraph, Alternative Spelling, Misspelling, Synonym Swap
& Rewrite, Homoglyph, Synonym, Misspelling, Whitespace \\

Prompt for generation & Simple text generation method & research-style, open-ended narrative, opinion-oriented prompts  &  Instruction-based, Task-oriented prompt\\

Human-written samples
& 7,043
& 4,000
& 83,900 \\

LLM-generated samples
& 29,627
& 12,000
& 503,400 \\

Adversarial samples
& --- 
& 12,000
& 419,500 \\

Total samples
& 36,670
& 28,000$^{*}$
& 1,006,800 \\

Primary focus
& Hindi news detection
& Multilingual detection and robustness under distribution shifts
& Large-scale Hindi AI-text detection with coverage of
Hindi-specific LLMs and adversarial robustness \\

\bottomrule
\end{tabular}%
}
\end{table*}

\section{Human Annotation and Quality Evaluation of the Proposed Dataset}

\subsection{Data Cleaning and Quality Control}
To mitigate potential biases arising from superficial textual cues that could bias AI-generated text detection, we apply consistent preprocessing to both human-written and LLM-generated Hindi texts.This includes removing repeated punctuation, bullet points, extraneous line breaks, HTML tags, and references, applying Unicode normalization, and removing sentences in languages other than Hindi. These operations aim to eliminate dataset-specific noise and formatting-related artifacts while preserving the original lexical content, word order, sentence structure, and semantics of the samples. The same preprocessing procedures are applied identically to both human-written and LLM-generated texts to avoid introducing or removing class-specific information. Consequently, the stylistic and semantic characteristics relevant to distinguishing AI-generated from human-written text remain intact. Formatting normalization is intended solely to improve textual consistency rather than alter authorship-related characteristics. These steps aim to reduce formatting-related artifacts while preserving the linguistic content of the samples.

Since sample provenance determines the original authorship labels, our quality-control procedure primarily assesses the linguistic quality and contextual appropriateness of LLM-generated texts. Specifically, we manually inspect 10--20 randomly selected generations per model and domain to evaluate fluency, coherence, plausibility, and relevance, and to identify systematic generation failures such as irrelevant responses, unintended language switching, excessive repetition, and malformed outputs. To minimise selection bias, we avoid filtering individual samples solely on the basis of their perceived human-likeness or detectability. Instead, we evaluate generation configurations for recurring quality issues and exclude configurations exhibiting systematic failures from subsequent text generation. To ensure consistent quality control across authorship categories, we apply common validation criteria to both human-written and LLM-generated Hindi texts, including language verification, duplicate detection, text-length validation, and the identification of empty or malformed samples. Additionally, we conduct source-specific quality assessments to account for differences in data provenance. For human-written texts, we examine source integrity and textual completeness; for LLM-generated texts, we manually inspect randomly selected samples to assess fluency, coherence, plausibility, and prompt relevance, and to identify systematic generation failures. This combined quality-control procedure aims to maintain dataset consistency without introducing unnecessary selection bias toward either authorship category. To avoid trivially short generated samples, we impose a minimum length threshold of approximately 1,000 characters on LLM-generated texts. This constraint encourages sufficiently developed outputs for meaningful evaluation. Overall, our quality-control procedure aims to preserve linguistic diversity while minimizing artifacts arising from inadequate prompting or generation failures.

\subsection{Human Annotations}
To further evaluate the quality and reliability of \textsc{SHAKTI}, we conduct a human annotation study on a randomly selected subset of eight hundred samples. Four annotators independently evaluate each sample using four criteria: \textbf{Fluency}, \textbf{Origin}, \textbf{Coherence}, and \textbf{Factuality}. All annotators are native Hindi speakers with strong linguistic competence. Two are graduate students in Computer Science with prior experience in Hindi NLP research, while the other two specialise in Hindi linguistics.
Fluency assesses grammatical correctness and readability, while Coherence measures the logical consistency and flow of ideas. Origin captures whether a text is perceived as human-written or LLM-generated, and Factuality evaluates the accuracy and relevance of its content. Fluency is rated on a three-point scale, Factuality on a five-point scale, and Origin and Coherence using binary labels. Appendix Table~\ref{tab:human_annotation} presents an annotation scheme for human evaluation metrics, including score labels and descriptions.  

\subsection{Inter-Annotator Agreement}
We assess annotation consistency using Krippendorff's alpha~\cite{krippendorff2011computing} and Fleiss' kappa~\cite{fleiss1971measuring}. Table~\ref{tab:agreement_metrics} presents the agreement scores across the four annotation dimensions.
The agreement for Origin, with a reported Fleiss' kappa of 0.54, highlights the difficulty of distinguishing human-written text from LLM-generated text based solely on textual characteristics. As LLMs increasingly produce fluent and coherent text, their outputs may closely resemble human writing, leading to differences in annotator judgments. Subjective interpretations of fluency, coherence, and factual accuracy may also contribute to disagreements. The relatively high agreement for Fluency, with a Fleiss' Kappa of 0.719, indicates substantial consistency among annotators in assessing the grammatical correctness and readability of the texts. This suggests that linguistic fluency is a comparatively well-defined evaluation dimension, allowing annotators to distinguish between fluent and grammatically deficient samples with relatively little disagreement. Similarly, the relatively high agreement for Factuality, with a Krippendorff's Alpha of 0.776, indicates considerable consistency among annotators in evaluating the perceived factual accuracy of textual content. This suggests that annotators generally share similar judgments regarding the plausibility and correctness of factual statements, although differences may arise when evaluating ambiguous claims, incomplete information, or subtle factual inconsistencies.

In summary, the inter-annotator agreement scores reported in Table~\ref{tab:agreement_metrics} provide complementary evidence regarding the consistency of human assessments of textual quality and perceived authorship beyond provenance-based labels. The results further highlight the challenges associated with human evaluation, particularly the variability in judgments when distinguishing human-written from LLM-generated Hindi texts.

\begin{table}[t]
\centering
\small
\caption{Krippendorff's Alpha and Fleiss' Kappa scores for reviews by metric, as mentioned in Appendix Table~\ref{tab:human_annotation}.}
\label{tab:agreement_metrics}

\renewcommand{\arraystretch}{1.15}
\setlength{\tabcolsep}{4pt}
\rowcolors{2}{gray!30}{white}

\resizebox{0.85\columnwidth}{!}{%
\begin{tabular}{lcccc}
\toprule
\rowcolor{blue!50!black}
\textcolor{white}{\textbf{Agreement Metric}} &
\textcolor{white}{\textbf{Fluency}} &
\textcolor{white}{\textbf{Origin}} &
\textcolor{white}{\textbf{Coherence}} &
\textcolor{white}{\textbf{Factuality}} \\
\midrule

Krippendorff's Alpha ($\alpha$) & 0.769 & 0.671 & 0.535 & 0.776 \\

Fleiss' Kappa ($\kappa$) & 0.719 & 0.543 & 0.456 & 0.543 \\

\bottomrule
\end{tabular}%
}
\end{table}

\section{Linguistic Feature-Based Comparison of Human vs. LLM-Generated Hindi Text}
\label{sec:feature_extraction}
In this study, we also investigate the differences between LLM-generated and human-written text using various linguistic features, aiming to understand how they differ in the context of Hindi language writing. We consider the following linguistic features to study differences between LLM-generated and human-written Hindi text. (i) Structural Feature, (ii) Lexical Diversity, (iii) Readability and Sentence Complexity, (iv) Indic Script and Semantic Distance and (v) Model Divergence between Human-Written and LLM-Generated Texts for Each LLM as noted in study\cite{kumarage2023stylometric, fontanarava2017feature}.

\subsection{Structural Comparison of LLM vs. Human Text}
\begin{table}[t]
\centering
\caption{Analysis of structural features in human-written and LLM-generated Hindi texts. \textbf{\# Characters} denotes the number of characters, \textbf{\# Words} the number of word tokens, \textbf{\# Sentences} and \textbf{\# Paragraphs} the number of sentences and paragraphs, respectively, and \textbf{Special Char} the number of special characters.}
\label{tab:structural_features}

\renewcommand{\arraystretch}{1.15}
\setlength{\tabcolsep}{4pt}
\rowcolors{2}{gray!30}{white}

\resizebox{\columnwidth}{!}{%
\begin{tabular}{lccccccccc}
\toprule
\rowcolor{blue!50!black}
\textcolor{white}{\textbf{Feature}} &
\textcolor{white}{\textbf{Human}} &
\textcolor{white}{\textbf{Gemma QA}} &
\textcolor{white}{\textbf{GPT}} &
\textcolor{white}{\textbf{LLaMA}} &
\textcolor{white}{\textbf{Mistral}} &
\textcolor{white}{\textbf{ChatGPT}} &
\textcolor{white}{\textbf{Gemini}} &
\textcolor{white}{\textbf{Aya}} &
\textcolor{white}{\textbf{Sarvam}} \\
\midrule

\# Characters & 170.73 & 825.08 & 740.44 & 4236.07 & \textbf{4572.91} & 895.42 & 1028.65 & 768.31 & 692.84 \\

\# Words & 29.19 & 135.91 & 127.18 & 709.69 & 772.70 & 148.26 & 169.83 & 126.74 & 114.52 \\

\# Sentences & 1.15 & 12.70 & 8.27 & 53.29 & 57.71 & 11.84 & 13.62 & 10.17 & 9.46 \\

\# Paragraphs & 1.00 & 22.56 & 6.74 & 42.05 & 47.24 & 7.31 & 8.26 & 6.12 & 5.48 \\

Special Char & \textbf{138.77} & 646.58 & 588.56 & 2857.06 & 3087.26 & 704.18 & 810.36 & 601.72 & 542.63 \\

Punctuation & 3.51 & 14.49 & 14.85 & 93.57 & 99.61 & 15.82 & 17.64 & 13.27 & 12.18 \\

\bottomrule
\end{tabular}%
}
\end{table}



Structural features capture measurable characteristics of a text, including its length, organisation, segmentation, and formatting, reflecting how the content is structured and presented. We analyse six structural features to quantify differences between
human-written and LLM-generated Hindi texts: character count, word
count, sentence count, paragraph count, special character count, and
punctuation count. The mathematical formulations and estimation procedures for these features are provided in Appendix Subsection~\ref{Appenix:structual}.

Table~\ref{tab:structural_features} presents a comparison between human-written text and text generated by various LLMs in terms of structural features. It is evident that the average character count for texts generated by Gemma, GPT, LLaMA, and Mistral is significantly higher than that of human-written text, indicating a tendency toward longer outputs. Similarly, the number of words, sentences, and paragraphs in LLM-generated text far exceeds that in human text, reflecting a more verbose and structured style. Additionally, the higher counts of special characters and punctuation in LLM-generated text suggest a more formal or technical tone, possibly influenced by the inclusion of structured content or coded information. LLMs generated Hindi text tends to be longer, more verbose, and structurally denser than human-written text, with higher counts of characters, words, sentences, and punctuation marks.

\begin{table}[t]
\centering
\caption{Comparative analysis of lexical diversity features in human-written and LLM-generated Hindi texts across different language models. \textbf{\# Unique Words} denotes the number of distinct word tokens, \textbf{TTR} (Type-Token Ratio) measures lexical diversity as the ratio of unique words to total words, and \textbf{Avg. Word Len.} represents the average number of characters per word. \textbf{Vowel Ratio} captures the proportion of vowels in the text, \textbf{Punctuation Var.} measures variation in punctuation usage, and \textbf{Repetition} quantifies repetitive patterns in the text. The reported values represent the average feature values for each text category or language model.}
\label{tab:lexical_features}

\renewcommand{\arraystretch}{1.15}
\setlength{\tabcolsep}{4pt}
\rowcolors{2}{gray!30}{white}

\resizebox{\columnwidth}{!}{%
\begin{tabular}{lccccccccc}
\toprule
\rowcolor{blue!50!black}
\textcolor{white}{\textbf{Feature}} &
\textcolor{white}{\textbf{Human}} &
\textcolor{white}{\textbf{Gemma QA}} &
\textcolor{white}{\textbf{GPT}} &
\textcolor{white}{\textbf{LLaMA}} &
\textcolor{white}{\textbf{Mistral}} &
\textcolor{white}{\textbf{ChatGPT}} &
\textcolor{white}{\textbf{Gemini}} &
\textcolor{white}{\textbf{Aya}} &
\textcolor{white}{\textbf{Sarvam}} \\
\midrule

\# Unique Words & 23.20 & 58.66 & 64.47 & 131.52 & 133.27 & 102.34 & 96.18 & 75.42 & 47.68 \\

TTR & 0.84 & 0.40 & 0.60 & 0.49 & 0.48 & 0.52 & 0.55 & 0.58 & 0.55 \\

Avg. Word Len. & 4.26 & 3.38 & 4.11 & 3.88 & 3.85 & 4.02 & 3.95 & 3.72 & 1.52 \\

Vowel Ratio & 0.00051 & 0.00089 & 0.00117 & 0.01105 & 0.01100 & 0.01012 & 0.00985 & 0.00872 & 0.06 \\

Punctuation Var. & 11.80 & 15.75 & 15.85 & 17.70 & 17.61 & 16.45 & 16.12 & 14.80 & 12.45 \\

Repetition & 0.11 & 0.14 & 0.16 & 0.15 & 0.15 & 0.13 & 0.14 & 0.12 & 0.23 \\

\bottomrule
\end{tabular}%
}
\end{table}


\subsection{Lexical Diversity in Human-Written and LLM-Generated Hindi Text}
Lexical diversity features capture measurable variations in vocabulary richness, word usage, lexical repetition, and character-level patterns, reflecting the diversity and distribution of linguistic elements within a text. We analyse six lexical and character-level features to investigate
differences in vocabulary richness, word length, character usage,
punctuation diversity, and repetition between human-written and
LLM-generated Hindi texts. The mathematical formulations and estimation procedures for these features are provided in Appendix Subsection~\ref{Appenix:lexical}.


Table~\ref{tab:lexical_features} compares human-written and LLM-generated Hindi texts in terms of lexical diversity and stylistic features. Overall, most LLMs produce more unique words than human authors, although their lower type-token ratios (TTRs) suggest a higher proportion of repeated words relative to their vocabulary size. LLM-generated texts generally exhibit shorter average word lengths and higher vowel ratios than human-written text, although the extent of these differences varies across models. Furthermore, LLM-generated texts show greater punctuation variety and slightly higher repetition rates in most cases, indicating differences in lexical choices and writing patterns. These findings highlight stylistic and linguistic distinctions between human-authored and LLM-generated Hindi texts, while also revealing substantial variation across language models.

\subsection{Evaluating Readability and Sentence Complexity}
\begin{table}[t]
\centering
\caption{Comparative analysis of readability and sentence complexity features in human-written and LLM-generated Hindi texts across different language models. \textbf{Avg. Word per Sent.} denotes the average number of words per sentence, while \textbf{Avg. Syll. per Word} represents the average number of syllables per word. \textbf{Flesch Reading} refers to the Flesch Reading Ease score, which estimates text readability, whereas the \textbf{Fog Index} and \textbf{SMOG Index} estimate textual complexity based on sentence length and word complexity. \textbf{Sentence Len. Var.} indicates the variability in sentence length. The reported values represent the average scores for each text category or language model.}
\label{tab:readability_features}

\renewcommand{\arraystretch}{1.15}
\setlength{\tabcolsep}{4pt}
\rowcolors{2}{gray!30}{white}

\resizebox{\columnwidth}{!}{%
\begin{tabular}{lccccccccc}
\toprule
\rowcolor{blue!50!black}
\textcolor{white}{\textbf{Feature}} &
\textcolor{white}{\textbf{Human}} &
\textcolor{white}{\textbf{Gemma QA}} &
\textcolor{white}{\textbf{GPT}} &
\textcolor{white}{\textbf{LLaMA}} &
\textcolor{white}{\textbf{Mistral}} &
\textcolor{white}{\textbf{ChatGPT}} &
\textcolor{white}{\textbf{Gemini}} &
\textcolor{white}{\textbf{Aya}} &
\textcolor{white}{\textbf{Sarvam}} \\
\midrule

Avg. Word per Sent. & 26.64 & 18.94 & 23.31 & 19.59 & 40.08 & 27.85 & 25.12 & 21.34 & 26.45 \\

Avg. Syll. per Word & 1.50 & 1.42 & 1.43 & 1.46 & 1.02 & 1.38 & 1.36 & 1.41 & 1.12 \\

Flesch Reading & 53.15 & 67.27 & 62.02 & 63.78 & 79.63 & 58.40 & 61.12 & 65.35 & 80.34 \\

Fog Index & 12.06 & 8.75 & 10.46 & 9.09 & 16.72 & 11.25 & 10.85 & 9.72 & 11.14 \\

SMOG Index & 7.38 & 5.90 & 6.64 & 6.27 & 6.22 & 6.91 & 6.58 & 6.12 & 4.55 \\

Sentence Len. Var. & 4.18 & 8.54 & 5.81 & 9.23 & 7.91 & 6.42 & 5.97 & 7.10 & 6.35 \\

\bottomrule
\end{tabular}%
}
\end{table}

Readability and sentence complexity features capture measurable characteristics of a text's linguistic accessibility and syntactic structure, reflecting the ease with which its content can be understood and the complexity of its sentence construction. Readability measures the ease with which a text can be understood,
whereas sentence complexity captures the structural characteristics
and variability of its sentences. We analyse six features to compare
human-written and LLM-generated Hindi texts: average words per
sentence, average syllables per word, Flesch Reading Ease, Gunning
Fog Index, SMOG Index, and sentence length variance. The mathematical formulations and estimation procedures for these features are provided in Appendix Subsection~\ref{Appenix:Readability}.


Table~\ref{tab:readability_features} compares human-written and LLM-generated Hindi texts in terms of readability and sentence complexity. Most LLMs exhibit higher Flesch Reading Ease scores and lower Fog and SMOG indices than human-written text, suggesting comparatively simpler linguistic structures, although this trend is not consistent across all models. Human-written text has the highest average syllables per word (1.50), indicating greater syllabic complexity. Sentence-length variability is also higher for all LLMs than for human-written text, with LLaMA and Gemma QA exhibiting the greatest variation, suggesting differences in sentence-length patterns. Mistral produces the longest sentences on average (40.08 words), whereas Sarvam achieves the highest Flesch Reading Ease score (80.34) and the lowest SMOG index (4.55). These differences highlight substantial variation in readability and sentence construction across models. Overall, the results demonstrate distinct linguistic and stylistic patterns in LLM-generated Hindi text compared with human-authored text.


\subsection{Indic Script and Semantic Distance-Based Evaluation}

\begin{table}[t]
\centering
\caption{Comparative analysis of Indic script and semantic distance features in human-written and LLM-generated Hindi texts across different language models. \textbf{Stopword Count} denotes the number of stopwords, while \textbf{Stopword Density} represents the proportion of stopwords relative to the total number of words. \textbf{Matra Count} measures the occurrence of Devanagari vowel signs in the text, and \textbf{BERT Dist.} denotes the BERT-based semantic distance, capturing semantic differences between text representations. The reported values represent the average feature values for each text category or language model.}
\label{tab:indic_semantic_features}

\renewcommand{\arraystretch}{1.15}
\setlength{\tabcolsep}{4pt}
\rowcolors{2}{gray!30}{white}

\resizebox{\columnwidth}{!}{%
\begin{tabular}{lccccccccc}
\toprule
\rowcolor{blue!50!black}
\textcolor{white}{\textbf{Feature}} &
\textcolor{white}{\textbf{Human}} &
\textcolor{white}{\textbf{Gemma QA}} &
\textcolor{white}{\textbf{GPT}} &
\textcolor{white}{\textbf{LLaMA}} &
\textcolor{white}{\textbf{Mistral}} &
\textcolor{white}{\textbf{ChatGPT}} &
\textcolor{white}{\textbf{Gemini}} &
\textcolor{white}{\textbf{Aya}} &
\textcolor{white}{\textbf{Sarvam}} \\
\midrule

Stopword Count & 6.60 & 31.29 & 23.44 & 128.68 & 136.94 & 88.52 & 75.63 & 52.74 & 18.35 \\

Stopword Density & 0.20 & 0.15 & 0.16 & 0.15 & 0.15 & 0.14 & 0.14 & 0.17 & 0.09 \\

Matra Count & 44.35 & 185.87 & 179.34 & 835.04 & 912.75 & 522.18 & 480.64 & 265.41 & 198.22 \\

BERT Dist. & 0.87 & 0.86 & 0.86 & 0.86 & 0.85 & 0.86 & 0.85 & 0.87 & 0.88 \\

\bottomrule
\end{tabular}%
}
\end{table}
We investigate four features to characterise differences between human-written and LLM-generated Hindi texts: stopword count, stopword density, Devanagari matra count, and BERT-based semantic distance. These features capture functional-word usage, script-specific characteristics, and semantic relationships between texts. The mathematical formulations and estimation procedures for these features are provided in Appendix Subsection~\ref{Appenix:indic}.

Table~\ref{tab:indic_semantic_features} compares human-written and LLM-generated Hindi texts in terms of stopword usage, Devanagari script characteristics, and semantic distance. All LLMs exhibit higher stopword counts than human-written text, with LLaMA and Mistral showing the highest values. However, stopword density is generally lower for LLM-generated texts, indicating that their higher stopword counts may primarily reflect differences in text length rather than a greater proportion of functional words. The matra count is also substantially higher for all LLMs, particularly LLaMA and Mistral, suggesting differences in the volume and composition of Devanagari text [33]. The BERT distance values remain relatively close across human-written and LLM-generated texts, ranging from 0.85 to 0.88. However, the interpretation of these values depends on the definition of the distance metric: lower values indicate greater semantic similarity if the measure represents distance, whereas higher values indicate greater similarity if it represents a similarity score. Overall, these findings highlight differences in textual length and script characteristics between human-written and LLM-generated Hindi texts, while motivating further investigation into their semantic relationships.

\begin{table}[t]
\centering
\caption{Comparative analysis of information-theoretic divergence between human-written and LLM-generated Hindi texts across different language models. $\mathrm{KL}(H \parallel M)$ measures the Kullback--Leibler divergence from the human lexical distribution $H$ to the model-generated distribution $M$, while $\mathrm{KL}(M \parallel H)$ measures the divergence in the reverse direction. Jensen--Shannon (JS) divergence provides a symmetric measure of distributional dissimilarity. Lower divergence values indicate greater similarity between human and model-generated lexical distributions.}
\label{tab:information_divergence}

\renewcommand{\arraystretch}{1.15}
\setlength{\tabcolsep}{4pt}
\rowcolors{2}{gray!30}{white}

\resizebox{\columnwidth}{!}{%
\begin{tabular}{lcccccccc}
\toprule
\rowcolor{blue!50!black}
\textcolor{white}{\textbf{Divergence Metric}} &
\textcolor{white}{\textbf{Gemma QA}} &
\textcolor{white}{\textbf{GPT}} &
\textcolor{white}{\textbf{LLaMA}} &
\textcolor{white}{\textbf{Mistral}} &
\textcolor{white}{\textbf{ChatGPT}} &
\textcolor{white}{\textbf{Gemini}} &
\textcolor{white}{\textbf{Aya}} &
\textcolor{white}{\textbf{Sarvam}} \\
\midrule

$\mathrm{KL}(H \parallel M)$
& 3.0207 & 2.1466 & 1.0611 & 3.2102 & 2.3846 & 2.6173 & 2.7935 & 2.4581 \\

$\mathrm{KL}(M \parallel H)$
& 2.3997 & 3.0345 & 2.4147 & 3.0036 & 2.7684 & 2.5831 & 2.9417 & 2.6259 \\

JS Divergence
& 0.3640 & 0.3729 & 0.3151 & 0.5567 & 0.3918 & 0.4086 & 0.4362 & 0.3785 \\

\bottomrule
\end{tabular}%
}
\end{table}

\subsection{Model Divergence between Human-Written and LLM-Generated Texts for Each LLM}
\label{app:kl}

$\mathrm{KL}(H\parallel M)$ quantifies the divergence of the model-generated distribution from the human-written distribution, with discrepancies
weighted according to the human-written distribution. Conversely, $\mathrm{KL}(M\parallel H)$ measures the reverse divergence, with discrepancies weighted
according to the model-generated distribution. The mathematical formulations and estimation procedures for these features are provided in Appendix Subsection~\ref{Appenix:divergence}.

 Table~\ref{tab:information_divergence} presents
information-theoretic divergence between
human-written and LLM-generated texts. Specifically, we compute the KL divergence in both directions and
Jensen--Shannon (JS) divergence to quantify
differences in their estimated lexical probability
distributions. Lower divergence (KL and JS ) values indicate
greater distributional similarity between
human-written and model-generated texts. Among the LLMs, LLaMA exhibits the lowest $\mathrm{KL}(H\parallel M)$ and
JS divergence , suggesting that its generated texts are lexically closer to human-written content. In contrast, Mistral records the highest $\mathrm{KL}(H\parallel M)$  and JS divergence, indicating greater lexical distributional differences. Notably,  GPT and Gemma fall between these extremes, showing moderate divergence. From such observation, we can conclude that certain LLMs generate texts whose lexical distributions closely resemble those of human-written content. In particular, LLaMA exhibits the strongest alignment under forward KL and JS divergence, indicating relatively smaller lexical distributional differences. Such similarities may reduce the effectiveness of detection approaches relying primarily on lexical distributional cues, although they do not necessarily imply that the generated texts are indistinguishable from human-written content.

Overall, the linguistic feature-based analysis reveals both similarities and differences between human-written and LLM-generated Hindi texts across structural, lexical, readability, and language-specific characteristics. The information-theoretic analysis further indicates that certain LLMs produce texts with lexical distributions closely aligned with human-written content, although the degree of alignment varies across generators. These findings highlight the limitations of relying solely on individual linguistic or lexical features to distinguish human-written from LLM-generated Hindi texts and motivate the investigation of complementary detection approaches that capture differences beyond surface-level linguistic patterns.

\section{Proposed Method}

We propose \textit{TRIDENT}, a round-trip translation (RTT)-based
framework for detecting large language model (LLM)-generated text. The primary motivation of our work is inspired by the hypothesis underlying RAIDAR~\cite{mao2024raidar} and SimLLM~\cite{nguyen2024simllm}, which posits that text generated by autoregressive language models retains structural patterns that another generative model is likely to recognize as high-quality text, resulting in less extensive rewriting. These studies demonstrate the potential of transformation-based signals for distinguishing human-written from LLM-generated content. However, rewriting differ from round-trip translation, which introduces cross-lingual transformations before reconstructing the text in its original language.

Motivated by this observation, we empirically examine the
transformation-induced representation shifts of human-written and
LLM-generated texts. As illustrated in Figure~\ref{fig:shift1_image},
the PCA visualization suggests that human-written samples generally
exhibit larger residual displacements than LLM-generated samples.
This preliminary observation motivates our hypothesis that
LLM-generated text may exhibit greater representational stability
under RTT, whereas human-written text may undergo greater
transformation-induced variation. We therefore investigate whether
these differences can be quantified using semantic similarity,
lexical overlap, and structural variation between the original
and round-trip translated texts, and whether they provide useful
signals for distinguishing human-written from LLM-generated content.

Appendix Figure~\ref{proposed_method} further illustrates the round-trip translation
(RTT) process for human-written and LLM-generated texts in Hindi,
Arabic, and Chinese, along with their corresponding cosine similarity
scores. The LLM-generated samples exhibit higher cosine similarity
than the human-written samples across all three languages, indicating
greater semantic preservation after RTT in these examples. The
observed differences motivate our hypothesis that human-written text
may undergo greater transformation-induced changes than LLM-generated
text. However, these examples are illustrative, and further quantitative
evaluation is required to establish the generalizability of this
observation across languages and domains.

Table~\ref{tab:information_divergence} results further support our RTT-based detection hypothesis. Models whose lexical distributions closely resemble human text tend to maintain stronger semantic consistency after round-trip translation, making them comparatively harder to distinguish. Conversely, models with higher divergence introduce larger semantic variations during translation, which increases detectable drift and facilitates more reliable separation from human-written text. 

Based on these observations, we hypothesize that LLM-generated text
exhibits greater semantic invariance under round-trip translation
(RTT) than human-written text. Specifically, LLM-generated text may
preserve its semantic content and linguistic structure more consistently
under cross-lingual transformations, whereas human-written text may
undergo greater semantic, lexical, and stylistic drift during translation
and reconstruction. This difference in transformation-induced variation
provides the motivation for using RTT-derived signals to distinguish
human-written from LLM-generated text.
\begin{figure}[t]
    \centering
\includegraphics[width=\linewidth]{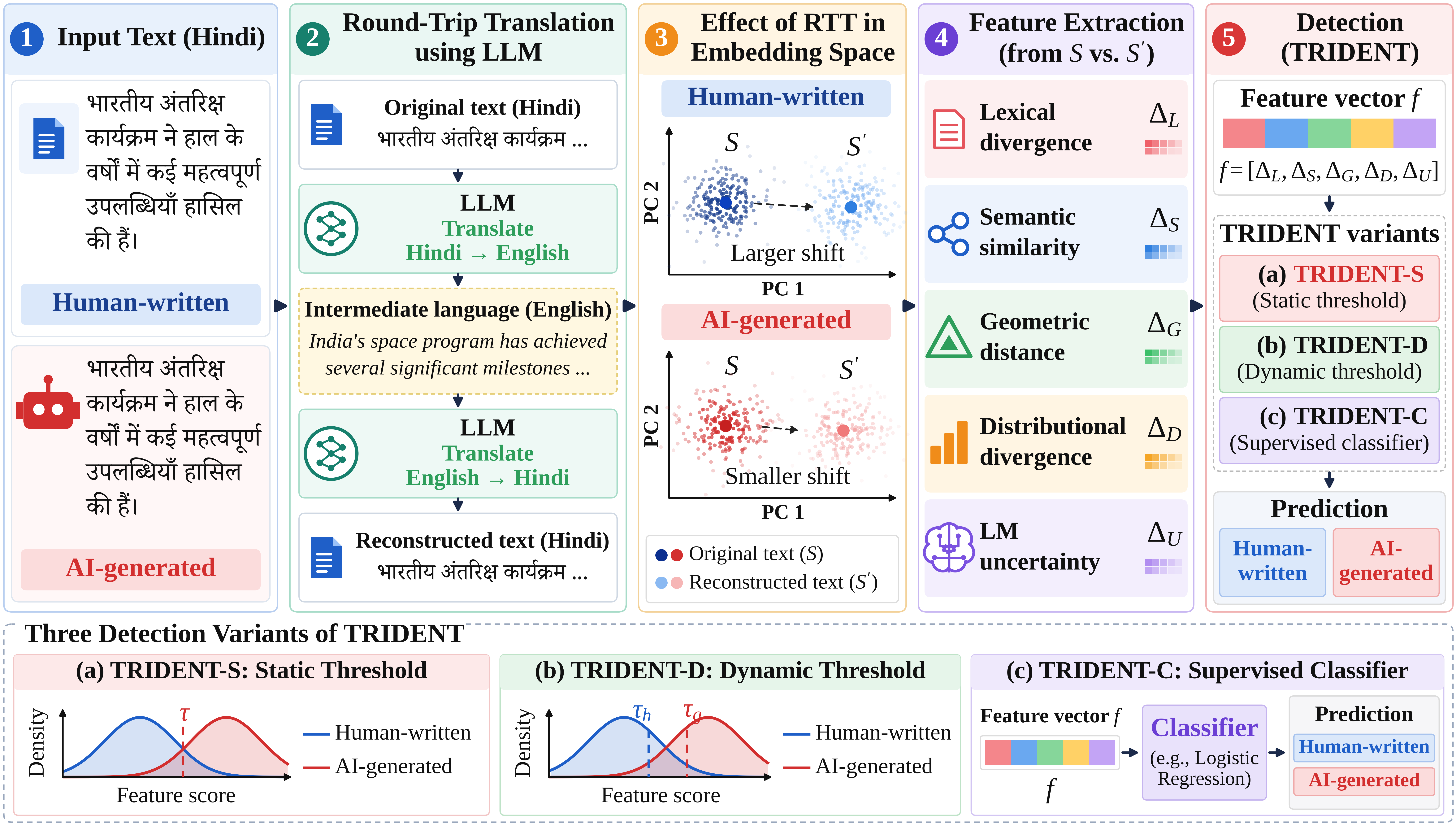}
    \caption{Overview of TRIDENT, a round-trip translation-based framework for AI-generated text detection. The input text is translated from the source language to an intermediate language and then back to the source language. The original and reconstructed texts are compared using lexical, semantic, geometric, distributional, and language-model-based features. These features support three detection variants: static-threshold classification (TRIDENT-S), dynamic-threshold classification (TRIDENT-D), and supervised classification (TRIDENT-C). }
    \label{fig:trident_framework}
\end{figure}
Similar observations regarding semantic drift are noted in a study~\cite{mao2024raidar}: neural language models tend to generate \textit{generic and self-consistent outputs} due to the dominance of high-likelihood token sequences during decoding. Consequently, LLM-generated text often behaves as a near \textit{fixed point} under repeated generative transformations such as translation or paraphrasing.

\begin{figure}[t]
    \centering
    \includegraphics[width=0.85\linewidth]{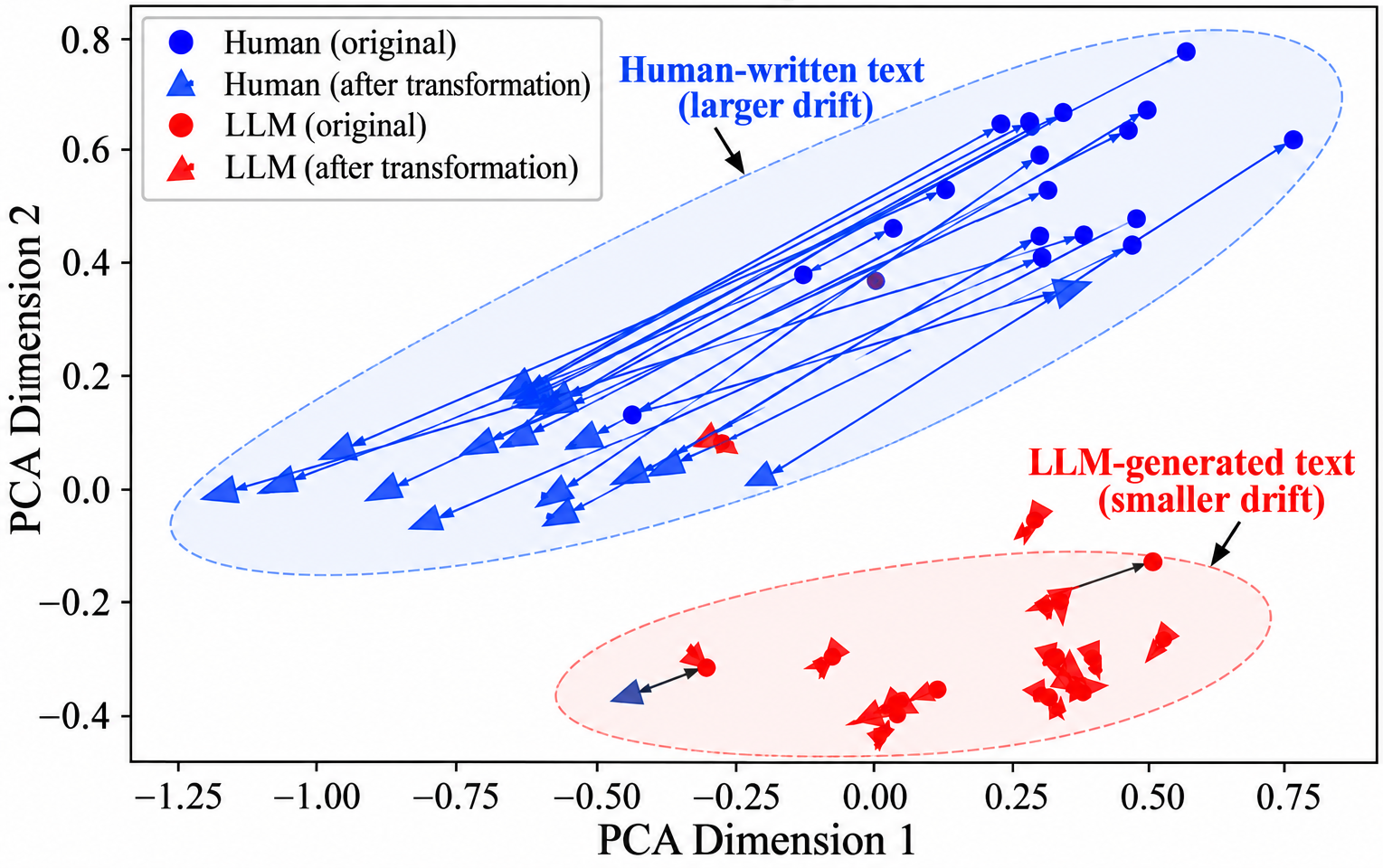} 
    \caption{ Two-dimensional PCA visualization of representation drift between original and round-trip-translated texts. Each point represents a text sample, and the corresponding arrow indicates the displacement induced by round-trip translation. Blue and red distinguish human-written and LLM-generated samples, respectively. The relative displacement patterns illustrate differences in transformation-induced variation between the two authorship classes. Residual drift between human-written and LLM-generated
    text in the two-dimensional PCA space. Each circle represents an
    original text sample, and the corresponding arrow indicates its
    displacement after the transformation. Blue and red denote
    human-written and LLM-generated samples, respectively. Human-written
    samples generally exhibit larger displacements, whereas LLM-generated
    samples tend to exhibit smaller displacements. This pattern suggests
    that the transformation affects the two text categories differently.}
    \label{fig:shift1_image}
    
\end{figure}

\subsection{Round-Trip Translation Operator} 

Let $\mathcal{S}$ denote an input sentence or paragraph drawn from a distribution $\mathcal{D}$ over a source language $L_1$. We define the round-trip translation operator $T_{\mathrm{RTT}}$ as a composition of two translation mappings performed using a translation model or large language model (LLM) parameterised by $\theta$.
\paragraph{Forward Translation ($L_1 \rightarrow L_2$):}
As shown in Figure~\ref{fig:first_image}, the original text $\mathcal{S}$ in language $L_1$ is translated into an intermediate language $L_2$ using the translation function $F_{\theta}$ such as $\mathcal{S}^{L_2} = F_{\theta}(\mathcal{S}; L_1 \rightarrow L_2)$, where $\mathcal{S}^{L_2}$ denotes the translated representation in the intermediate language.
\paragraph{Backward Translation ($L_2 \rightarrow L_1$)}
The intermediate text $\mathcal{S}^{L_2}$ is then translated back to the source language $L_1$: $\mathcal{S}^{L_1'} = F_{\theta}(\mathcal{S}^{L_2}; L_2 \rightarrow L_1)$, where $\mathcal{S}^{L_1'}$ represents the reconstructed text obtained after round-trip translation(Figure~\ref{fig:first_image}).The complete round-trip transformation can therefore be written as the composition described in Equation~\ref{equation1} below. 

\begin{equation}  
\label{equation1}
\small
    \mathcal{S}^{L_1'} =  T_{\mathrm{RTT}}(\mathcal{S}) = F_{\theta}^{L_2 \rightarrow L_1}
\left(
F_{\theta}^{L_1 \rightarrow L_2}(\mathcal{S})
\right)  
\end{equation} 
This transformation acts as a \textit{controlled perturbation operator} that approximately preserves the semantic content of the original text while introducing lexical, syntactic, and stylistic variations induced by the translation process. Such perturbations allow us to analyse the stability of textual representations under generative transformations, which form the basis of our TRIDENT framework.

\subsection{Embedding Representation of Original and RTT Text}
\label{subsec:feature_esti}
To systematically quantify the drift induced by the round-trip translation (RTT) process, we encode both the original text $\mathcal{S}$ and its RTT reconstruction $\mathcal{S}^{L_1'}$ into continuous vector representations using a sentence encoder. Let $\mathbf{s} = E(\mathcal{S})$, $\mathbf{s}^{r} = E(\mathcal{S}^{L_1'})$ denote the embedding representations of the original and RTT-transformed texts, respectively, where $E(\cdot)$ represents a embedding model that maps a sentence into a $d$-dimensional semantic vector space. These embeddings capture semantic and contextual information across languages in a shared representation space, enabling direct comparison between the original and reconstructed texts. Intuitively, larger divergence between $\mathbf{s}$ and $\mathbf{s}^{r}$ indicates stronger transformation induced by RTT. To quantify this divergence, we compute several complementary metrics between $\mathcal{S}$ and its RTT-transformed counterpart $\mathcal{S}^{L_1'}$:
\begin{itemize} 
\item \textbf{Lexical divergence ($L_d$):} Edit distance between the token sequences of $\mathcal{S}$ and $\mathcal{S}^{L_1'}$. 
\item \textbf{Semantic similarity ($S_c$):} Cosine similarity between the embeddings $\mathbf{s}$ and $\mathbf{s}^{r}$. 
\item \textbf{Angular distance ($A_d$):} Angle between $\mathbf{s}$ and $\mathbf{s}^{r}$ computed as $\arccos(S_c)$. 
\item \textbf{Geometric distance ($E_d$):} Euclidean distance $\|\mathbf{s} - \mathbf{s}^{r}\|_2$ in the embedding space. 
\item \textbf{Distributional divergence ($D_{KL}, D_{JS}$):} KL divergence and Jensen–Shannon distance between token probability distributions. 
\item \textbf{Language model uncertainty ($P_d$):} Perplexity difference $|\text{PPL}(\mathcal{S}) - \text{PPL}(\mathcal{S}^{L_1'})|$. 
\end{itemize}
These metrics collectively provide a multi-dimensional characterization of RTT-induced drift, capturing lexical changes, semantic shifts, and probabilistic differences between the original and reconstructed texts. The resulting feature set forms the basis of our detection framework for distinguishing human-authored text from LLM-generated text. 
\subsection{Classification Model}
We concatenate all RTT-induced drift features into a unified feature representation. Let
\begin{equation}
\label{equation_f}
    \mathbf{f} = [L_d, S_c, A_d, E_d, D_{KL}, D_{JS}, P_d] 
\end{equation}  $\mathbf{f}$ denote the feature vector capturing lexical, semantic, geometric, and distributional divergence between the original text $\mathcal{S}$ and its RTT-transformed counterpart $\mathcal{S}^{L_1'}$.  Building on these representations, we introduce three
complementary detection approaches that characterize
translation-induced semantic drift by comparing the
embeddings of the original text and its RTT-transformed
counterpart, denoted by $(\mathbf{s}, \mathbf{s}^{r})$.


\subsubsection{\textbf{TRIDENT-S} (Static Threshold-based Classification):} 
The first variant adopts a simple zero-shot formulation based on cosine similarity between the embeddings of the original text and its round-trip counterpart. The motivation behind this approach is twofold: (i) it offers a fully model-agnostic and training-free solution, and (ii) it enables direct deployment in real-world settings where labelled Hindi data or access to model internals is unavailable. By identifying a decision threshold on RTT similarity, this method leverages semantic stability as a direct indicator of machine-generated text. We quantify semantic preservation using cosine similarity $S_c$ between $\mathbf{s}$ and $\mathbf{s}^{r}$. In \textbf{TRIDENT-S}, if $S_c < \alpha$, the text is classified as human-written; otherwise, it is considered LLM-generated. The decision threshold $\alpha$ for LLM-generated text classification is determined using Receiver Operating Characteristic (ROC) analysis, as detailed in Figure~\ref{tab:threshold}.
To determine an effective decision threshold value $\alpha$ for distinguishing between human-written and LLM-generated Hindi text, we simulated a dataset consisting of 200 samples, 100 human-written and 100 LLM-generated texts. Cosine similarity scores were computed for each sample, with human-written texts exhibiting higher average similarity compared to LLM-generated texts.  We then evaluated the performance of different thresholds using a Receiver Operating Characteristic (ROC) curve and calculated the corresponding True Positive Rate (TPR) and False Positive Rate (FPR). Rather than selecting the threshold that maximises Youden’s J statistic. The selected threshold was highlighted on the ROC curve and used in subsequent evaluation in Figure~\ref{tab:threshold}. The key steps involved in the TRIDENT-S classification procedure are as follows.

\begin{enumerate}
    \item \textbf{Compute Similarity Scores:} Compute the cosine similarity score $S_c$ between
the original text embedding $\mathbf{s}$ and its
round-trip translated counterpart $\mathbf{s}^{r}$
for each sample in the validation dataset.

    \item \textbf{Evaluate Candidate Thresholds:}
    Vary the decision threshold $\alpha$ over the
    observed similarity range. For each threshold,
    classify a sample as LLM-generated if its
    similarity score is greater than or equal to
    $\alpha$; otherwise, classify it as human-written.

    \item \textbf{Construct the ROC Curve:}
    For each threshold, compute the True Positive
    Rate (TPR) and False Positive Rate (FPR).
    Construct the ROC curve by plotting FPR on the
    horizontal axis and TPR on the vertical axis.

    \item \textbf{Select the Optimal Threshold:}
    Determine the optimal threshold $\alpha^*$
    by maximizing Youden's $J$ statistic,
    $J(\alpha)=\mathrm{TPR}(\alpha)-\mathrm{FPR}(\alpha)$.

    \item \textbf{Perform Final Classification:}
   Apply the threshold $\alpha^*$ selected for TRIDENT-S to the held-out test dataset without further adjustment to evaluate detection performance.
\end{enumerate}

\subsubsection{\textbf{TRIDENT-D} (Dynamic Threshold-Based Classification):} While TRIDENT-S (static threshold) classification relies on a globally tuned decision boundary $\alpha$, it may not generalise well across datasets or encoder variants. To address this limitation, we adopt a TRIDENT-D, dynamic threshold adaptation strategy based on unsupervised Gaussian Mixture Modelling (GMM). The motivation for TRIDENT-D is to allow the decision boundary to automatically adapt to the intrinsic similarity distribution of a given dataset. By modelling RTT similarity scores as a mixture of human-written and LLM-generated text distributions, this approach eliminates the need for manual threshold tuning and enhances robustness under domain shifts and unseen model conditions.

Given an input text $\mathcal{S}$ and its round-trip-translated counterpart $\mathcal{S}^{L_1'}$, a pretrained sentence encoder generates their representations, which are used to compute the cosine similarity score $S_c$. These scores are collected from an unlabeled development set and modeled using a two-component Gaussian mixture model:
\begin{equation}
\label{equation:3}
p(S_c)=\sum_{k=1}^{2}\pi_k
\mathcal{N}(S_c\mid\mu_k,\sigma_k^2),
\end{equation}
where $\mathcal{N}(S_c\mid\mu_k,\sigma_k^2)$ denotes a univariate Gaussian distribution with mean $\mu_k$ and variance $\sigma_k^2$, and $\pi_k$ is the mixing coefficient of the $k$-th component, satisfying $\pi_k\geq 0$ and $\sum_{k=1}^{2}\pi_k=1$. The mixture parameters are estimated using the Expectation--Maximization (EM) algorithm.

After fitting the GMM, the component means are ordered such that $\mu_{\text{low}}<\mu_{\text{high}}$. Following the assumption that LLM-generated texts tend to exhibit lower RTT similarity scores than human-written texts, we define the adaptive threshold as the midpoint between the two component means:
\begin{equation}
\label{eq}
\alpha_{\text{GMM}}=
\frac{\mu_{\text{low}}+\mu_{\text{high}}}{2}.
\end{equation}
For a test instance with similarity score $S_c$, the predicted label is determined as
\begin{equation}
\hat{y}=
\begin{cases}
\text{LLM generated}, & S_c\geq\alpha_{\text{GMM}},  \\
\text{Human}, & S_c<\alpha_{\text{GMM}}.
\end{cases}
\end{equation}

In \textbf{TRIDENT-D}, The estimated threshold is held fixed during test evaluation. Consequently, the method does not require ROC-based threshold optimization or labeled test data. By adapting the decision boundary to the similarity-score distribution observed in the development set, TRIDENT-D provides a simple, training-free alternative to static thresholding.


\subsubsection{\textbf{TRIDENT-C} (Logistic Regression with Multi-Distance RTT Features):} 
While cosine similarity captures coarse semantic stability, it may not fully reflect finer lexical and distributional differences introduced by RTT. To address this limitation, 
We construct a richer feature representation $\mathbf{f}$ from the RTT-induced drift features estimated in Section~\ref{subsec:feature_esti} and concatenate them into a single feature vector, as defined in Equation~\ref{equation_f}. TRIDENT-C uses this feature vector to train a logistic regression classifier that estimates the probability of a text being LLM-generated, as defined in Equation~\ref {equation5}.
\begin{equation}  
\label{equation5}
    \hat{y} = \sigma(\mathbf{w}^\top \mathbf{f} + b), 
\end{equation}
where $\sigma(\cdot)$ denotes the sigmoid function. The model is optimised using binary cross-entropy loss, with gradients backpropagated jointly through both the encoder and classifier parameters. This formulation enables the model to capture complementary RTT-induced divergence signals, resulting in a more expressive and discriminative detection framework.



\begin{table*}[t]
\centering
\footnotesize
\caption{Dataset statistics, train, development, test splits, and
decision thresholds used in the TRIDENT-S experiments.}
\label{tab:dataset_statistics}

\renewcommand{\arraystretch}{1.2}
\setlength{\tabcolsep}{5pt}
\rowcolors{2}{gray!30}{white}

\resizebox{\textwidth}{!}{%
\begin{tabular}{lrrrrrrrrr}
\toprule
\rowcolor{blue!50!black}
\textcolor{white}{\textbf{Metric}} &
\textcolor{white}{\textbf{SHAKTI}} &
\textcolor{white}{\textbf{AGhi}} &
\textcolor{white}{\textbf{Urdu}} &
\textcolor{white}{\textbf{Arabic}} &
\textcolor{white}{\textbf{Korean}} &
\textcolor{white}{\textbf{Chinese}} &
\textcolor{white}{\textbf{Russian}} &
\textcolor{white}{\textbf{Bulgarian}} &
\textcolor{white}{\textbf{English}} \\
\midrule

Train
& 352k & 22k & 600 & 4.8k & 1.2k & 3.6k & 3.6k & 3.6k & 16k \\

Dev
& 117k & 7k & 200 & 1.6k & 400 & 1.2k & 1.2k & 1.2k & 4k \\

Test
& 117k & 7k & 200 & 1.6k & 400 & 1.2k & 1.2k & 1.2k & 4k \\

\midrule
\rowcolor{gray!50}
\textbf{Threshold}
& \textbf{0.875}
& \textbf{0.861}
& \textbf{0.758}
& \textbf{0.856}
& \textbf{0.802}
& \textbf{0.831}
& \textbf{0.689}
& \textbf{0.806}
& \textbf{0.816} \\

\bottomrule
\end{tabular}%
}
\end{table*}


\section{Experimental Setup}
\label{sec}

This section describes the baseline methods, experimental datasets, evaluation metrics, and evaluation protocol used to assess the effectiveness of the proposed TRIDENT framework. We compare TRIDENT against representative AI-generated text detection methods spanning statistical, perturbation-based, rewriting-based, regeneration-based, and supervised classification approaches. This selection enables us to investigate the relative effectiveness of different detection strategies across multiple text genres and datasets.

\subsection{Baseline Methods}
\label{subsec}

We compare the proposed TRIDENT framework against representative methods for detecting LLM-generated text. The baselines cover statistical, perturbation-based, rewriting-based, and supervised multilingual detection approaches. This selection allows us to evaluate TRIDENT against methods that exploit different signals, including language-model probabilities, text transformations, and contextual representations.

\noindent\textbf{Statistical and perturbation-based methods.}
We consider Log-Likelihood, Rank, Log-Rank, and Entropy~\cite{gehrmann2019gltr}~\footnote{\href{https://github.com/HendrikStrobelt/detecting-fake-text}{https://github.com/HendrikStrobelt/detecting-fake-text}}, which exploit token-level probabilities and predictive uncertainty. We also include DetectGPT~\cite{mitchell2023detectgpt}~\footnote{\href{https://github.com/eric-mitchell/detect-gpt}{https://github.com/eric-mitchell/detect-gpt}}, Fast-DetectGPT~\cite{bao2024fast}~\footnote{\href{https://github.com/baoguangsheng/fast-detect-gpt}{https://github.com/baoguangsheng/fast-detect-gpt}}, and DetectLLM-NPR~\cite{su2023detectllm} (NPR)~\footnote{\href{https://github.com/mbzuai-nlp/DetectLLM}{https://github.com/mbzuai-nlp/DetectLLM}}, which use probability curvature or perturbation-based statistics for zero-shot detection.

\noindent\textbf{Rewriting-based and additional detection methods.}
We include RAIDAR~\cite{mao2024raidar}~\footnote{\href{https://github.com/cvlab-columbia/raidarllmdetect}{https://github.com/cvlab-columbia/raidarllmdetect}} and LLaMA-L2R~\cite{hao-etal-2025-learning}, which exploits language-model rewriting behavior, and SimLLM~\cite{nguyen2024simllm}~\footnote{\href{https://github.com/quocnsh/SimLLM}{https://github.com/quocnsh/SimLLM}}, which uses similarity between original and regenerated text. We also consider J-Guard~\cite{kumarage2023j}~\footnote{\href{https://github.com/TSKumarage/J-Guard}{https://github.com/TSKumarage/J-Guard}}, a journalism-guided framework for adversarially robust detection of AI-generated news, and BiScope~\cite{guo2024biscope}~\footnote{\href{https://github.com/MarkGHX/BiScope}{https://github.com/MarkGHX/BiScope}}, which exploits token-level prediction statistics to identify memorization patterns.

\noindent\textbf{Supervised multilingual baselines.}
We use XLM-RoBERTa~\cite{conneau2019cross} as a supervised multilingual baseline. We additionally evaluate a feature-augmented variant that combines contextual embeddings with the 22 textual features described in Section~\ref{sec}. Let $\mathbf{h}_i\in\mathbb{R}^{768}$ denote the classification-token representation of input text $S_i$, and let $\mathbf{f}_i\in\mathbb{R}^{22}$ denote its extracted feature vector. The combined representation is obtained by concatenation:

\begin{equation}
\mathbf{z}_i = [\mathbf{h}_i;\mathbf{f}_i]
\in \mathbb{R}^{790}.
\end{equation}

The combined representation is passed to a feedforward classifier to distinguish human-written text from LLM-generated text. This variant allows us to investigate whether incorporating explicit textual features improves detection performance over contextual representations alone.

\noindent\textbf{Proposed TRIDENT variants.}
We evaluate three variants of the proposed TRIDENT framework: \textbf{TRIDENT-S}, which employs static-threshold classification; \textbf{TRIDENT-D}, which uses dynamic-threshold classification; and \textbf{TRIDENT-C}, which applies logistic regression to multi-distance round-trip translation (RTT) features. These variants enable us to compare fixed and adaptive threshold-based decision rules with a learned classification boundary based on RTT-induced textual variations.





\subsection{Evaluation Datasets}
\label{subsec}

We evaluate the proposed framework and Baselines method across datasets covering five languages: Hindi, Urdu, Arabic, Korean, and English. The primary evaluation uses our proposed SHAKTI dataset, which contains Hindi text from four genres: QA, News, Wiki, and Essay. We additionally use the existing Hindi dataset $AG_{\text{hi}}$~\cite{kavathekar2024counter}, focusing on its News subset.

To investigate multilingual applicability, we evaluate our framework on the Urdu dataset~\cite{ali2025hlu}, Arabic dataset~\cite{al2025arabic}, and Korean dataset~\cite{park2025detecting}. We also report results on the Bulgarian and Chinese subsets of M4 and the English RAID dataset. These evaluations provide complementary evidence of performance across languages and datasets with different linguistic and domain characteristics. Table~\ref{tab:dataset_statistics} summarizes the key characteristics of the datasets used in this study.




\subsection{Evaluation Metrics}
\label{subsec}

We evaluate detection performance using accuracy (Acc) and class-wise F1 scores for human-written (H) and LLM-generated (AI) texts. Accuracy measures the proportion of correctly classified samples among all evaluated samples. The class-wise F1 score is the harmonic mean of precision and recall calculated separately for each class.

We denote the F1 scores for the human-written and AI-generated classes as F1-H and F1-AI, respectively. Reporting these scores alongside accuracy provides a more detailed assessment of classification performance by showing how effectively each method identifies the two classes. The same metric definitions and class-label conventions are applied across the evaluated methods.

\subsection{Implementation Details and Experimental Protocol}
\label{subsec}

Appendix Table~\ref{tab:hyperparams} summarizes the hyperparameters used for RTT-based detection, threshold estimation, and the associated classification components. The similarity threshold for static stylistic-drift detection is selected using the validation set, with the search range beginning at 0.875. The final threshold is then applied to the test data without using test labels for threshold selection.

TRIDENT comprises three variants: (i) TRIDENT-S, which selects an optimal static decision threshold using labelled validation data and keeps it fixed during test data; (ii) TRIDENT-D estimates a threshold using a two-component Gaussian Mixture Model (GMM) on unlabeled validation scores and applies it unchanged to the test set; and (iii) TRIDENT-C trains a logistic regression classifier on labeled RTT-derived features using the training set and evaluates it on the test set. TRIDENT-S and TRIDENT-D independently estimate language-specific thresholds using their respective validation partitions, whereas TRIDENT-C is trained independently using the corresponding language-specific training partitions. The detailed statistics of the training, validation, test sets and threshold of TRIDENT-S are presented in Table~\ref{tab:dataset_statistics}. The language-specific training partitions are used to train TRIDENT-C, while the corresponding validation partitions are used to estimate thresholds for TRIDENT-S and TRIDENT-D. The resulting thresholds and trained classifiers remain fixed, with the test partitions reserved exclusively for the final evaluation of all three variants.

All experiments are conducted on NVIDIA A100 GPUs with 40 GB of memory. Fixed random seeds are used to improve reproducibility. For supervised methods, training, validation, and test partitions are kept separate, and model selection is performed without access to test labels. For zero-shot methods, the scoring procedure and underlying language-model configuration are specified according to the original method. Any adaptations needed to support the evaluated languages, datasets, or input formats are documented with the corresponding implementation.

\begin{table*}[t]
\centering
\small
\caption{Comparative analysis of accuracy and class-wise F1 scores for detecting human-written (H) and LLM-generated (AI) Hindi texts across different genres of the SHAKTI dataset and the existing $AG_{\text{hi}}$ dataset~\cite{kavathekar2024counter}. Results are reported for statistical, zero-shot, and supervised detection methods, including three TRIDENT variants: TRIDENT-S, TRIDENT-D, and TRIDENT-C. \textbf{Bold} values indicate the best performance among the TRIDENT variants for each dataset genre and metric, while \textcolor{blue}{blue} values highlight the best performance among all evaluated methods.}
\label{tab:llm_detection_results}
\renewcommand{\arraystretch}{1.15}
\setlength{\tabcolsep}{3pt}
\rowcolors{2}{gray!30}{white}

\resizebox{\textwidth}{!}{%
\begin{tabular}{lccccccccccccccc}
\toprule
\rowcolor{blue!50!black}
\textcolor{white}{\textbf{Model}}
& \multicolumn{12}{c}{\textcolor{white}{\textbf{SHAKTI}}}
& \multicolumn{3}{c}{\textcolor{white}{\textbf{$AG_{\text{hi}}$}}} \\
\cmidrule(lr){2-13}
\cmidrule(lr){14-16}
\rowcolor{blue!50!black}
\textcolor{white}{\textbf{Genre}}
& \multicolumn{3}{c}{\textcolor{white}{\textbf{QA}}}
& \multicolumn{3}{c}{\textcolor{white}{\textbf{News}}}
& \multicolumn{3}{c}{\textcolor{white}{\textbf{Wiki}}}
& \multicolumn{3}{c}{\textcolor{white}{\textbf{Essay}}}
& \multicolumn{3}{c}{\textcolor{white}{\textbf{News}}} \\

\rowcolor{blue!50!black}
& \textcolor{white}{\textbf{Acc}} & \textcolor{white}{\textbf{F1-H}} & \textcolor{white}{\textbf{F1-AI}}
& \textcolor{white}{\textbf{Acc}} & \textcolor{white}{\textbf{F1-H}} & \textcolor{white}{\textbf{F1-AI}}
& \textcolor{white}{\textbf{Acc}} & \textcolor{white}{\textbf{F1-H}} & \textcolor{white}{\textbf{F1-AI}}
& \textcolor{white}{\textbf{Acc}} & \textcolor{white}{\textbf{F1-H}} & \textcolor{white}{\textbf{F1-AI}}
& \textcolor{white}{\textbf{Acc}} & \textcolor{white}{\textbf{F1-H}} & \textcolor{white}{\textbf{F1-AI}} \\
\midrule

Log-Likelihood~\cite{gehrmann2019gltr}
& 0.29 & 0.28 & 0.31 & 0.29 & 0.26 & 0.32 & 0.32 & 0.30 & 0.34 & 0.27 & 0.25 & 0.29 & 0.30 & 0.28 & 0.31 \\

Rank~\cite{gehrmann2019gltr}
& 0.43 & 0.41 & 0.44 & 0.45 & 0.44 & 0.46 & 0.44 & 0.43 & 0.45 & 0.42 & 0.41 & 0.43 & 0.45 & 0.44 & 0.47 \\

Log-Rank~\cite{gehrmann2019gltr}
& 0.45 & 0.44 & 0.47 & 0.47 & 0.46 & 0.49 & 0.47 & 0.46 & 0.48 & 0.45 & 0.44 & 0.47 & 0.47 & 0.45 & 0.50 \\

Entropy~\cite{gehrmann2019gltr}
& 0.49 & 0.48 & 0.50 & 0.50 & 0.49 & 0.52 & 0.51 & 0.50 & 0.51 & 0.48 & 0.47 & 0.49 & 0.51 & 0.50 & 0.53 \\

DetectLLM-NPR~\cite{su2023detectllm}
& 0.64 & 0.65 & 0.64 & 0.60 & 0.59 & 0.61 & 0.63 & 0.70 & 0.52 & 0.59 & 0.57 & 0.61 & 0.61 & 0.59 & 0.64 \\

DetectGPT~\cite{mitchell2023detectgpt}
& 0.60 & 0.59 & 0.61 & 0.63 & 0.62 & 0.64& 0.61 & 0.60 & 0.62 & 0.63 & 0.61 & 0.66 & 0.64 & 0.62 & 0.66 \\

Fast-DetectGPT~\cite{bao2024fast}
& 0.78 & 0.80 & 0.77 & 0.77 & 0.74 & 0.80 & 0.79 & 0.77 & 0.81 & 0.77 & 0.75 & 0.80 & 0.82 & \textbf{0.81} & \textbf{0.84} \\

J-Guard~\cite{kumarage2023j}
& 0.79 & 0.81 & 0.77 & \textbf{0.82} & \textbf{0.83} & 0.81 & 0.78 & 0.77 & 0.79 & \textbf{0.81} & \textbf{0.80} & \textbf{0.83} & 0.80 & \textbf{0.81} & 0.78 \\

Biscope~\cite{guo2024biscope}
& \textbf{0.84} & \textcolor{blue}{0.86} & \textbf{0.83} & 0.81 & 0.79 & 0.82 & 0.80 & 0.78 & 0.81 & 0.77 & 0.75 & 0.78 & 0.82 & 0.80 & 0.83 \\

RAIDAR~\cite{mao2024raidar}
& 0.76 & 0.75 & 0.77 & 0.78 & 0.77 & 0.80 & 0.77 & 0.76 & 0.78 & 0.75 & 0.74 & 0.76 & 0.76 & 0.76 & 0.77 \\

SimLLM~\cite{nguyen2024simllm}
& 0.81 & 0.82 & 0.79 & 0.82 & 0.80 & \textbf{0.83} & \textbf{0.81} & 0.79 & \textbf{0.82} & 0.79 & 0.77 & 0.80 & \textbf{0.83} & \textbf{0.81} & \textbf{0.84} \\

LLaMA-L2R~\cite{hao-etal-2025-learning}
& 0.79 & 0.77 & 0.80 & 0.79 & 0.75 & 0.82 & 0.79 & \textcolor{blue}{0.80} & 0.79 & 0.73 & 0.75 & 0.71 & 0.81 & 0.79 & 0.82 \\

\midrule

XLM-RoBERTa~\cite{conneau2019cross}
& 0.76 & 0.64 & 0.82 & 0.78 & 0.67 & 0.83 & 0.80 & 0.75 & {0.83} & 0.78 & 0.66 & 0.84 & 0.79 & 0.70 & 0.84 \\

XLM-RoBERTa + feature
& \textbf{0.80} & \textbf{0.71} & \textbf{0.85} & \textbf{0.82} & \textbf{0.77} & \textbf{0.85} & \textbf{0.83} & \textbf{0.74} & {0.87} & \textbf{0.83} & \textbf{0.71} & \textbf{0.88} & \textbf{0.85} & \textbf{0.74} & \textbf{0.89} \\

\midrule
\rowcolor{gray!50}
\textbf{TRIDENT-S (Static Threshold)}
& \textcolor{blue}{0.89} & 0.83 & \textcolor{blue}{0.92}
& 0.89 & \textcolor{blue}{0.85} & 0.91
& 0.84 & \textcolor{blue}{0.80} & 0.87
& \textcolor{blue}{0.88} & 0.79 & \textcolor{blue}{0.92}
& \textcolor{blue}{0.90} & \textcolor{blue}{0.86} & 0.91 \\

\rowcolor{gray!30}
\textbf{TRIDENT-D (Dynamic Threshold)}
& 0.85 & \textbf{0.84} & 0.86
& 0.87 & 0.83 & 0.89
& \textcolor{blue}{0.85} & 0.77 & \textcolor{blue}{0.89}
& 0.87 & \textcolor{blue}{0.83} & 0.89
& \textcolor{blue}{0.90} & 0.85 & \textcolor{blue}{0.92} \\

\rowcolor{gray!50}
\textbf{TRIDENT-C (Logistic Regression)}
& 0.88 & 0.83 & 0.91
& \textcolor{blue}{0.90} & 0.84 & \textcolor{blue}{0.93}
& 0.84 & 0.78 & \textcolor{blue}{0.88}
& 0.88 & 0.77 & \textcolor{blue}{0.92}
& 0.88 & \textcolor{blue}{0.86} & 0.90 \\

\bottomrule
\end{tabular}%
}
\end{table*}

\begin{figure}[t]
    \centering
    \includegraphics[width=0.85\linewidth]{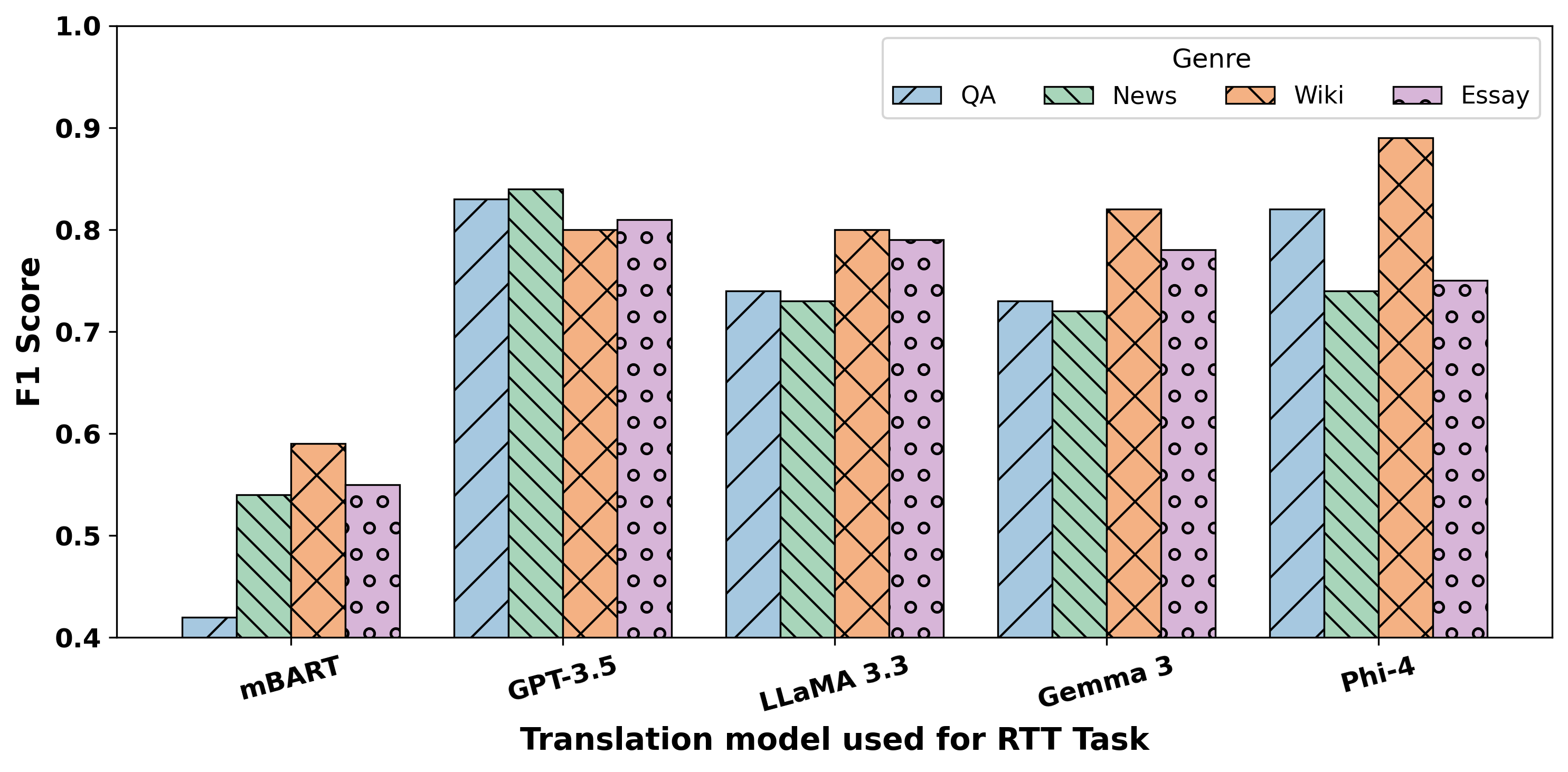}
    \caption{\small F1- score comparison of TRIDENT-S(Static Threshold) performance using mBART versus LLM-based round-trip translation models.}
    \label{fig:mbart}
\end{figure}

\begin{table*}[t]
\centering
\small
\caption{Comparative analysis of multilingual TRIDENT performance using
different round-trip translation models across Urdu, Arabic, Korean,
Russian, Bulgarian, Chinese, and English (RAID) evaluation datasets.
Accuracy (Acc) and F1 scores are reported for the TRIDENT-S,
TRIDENT-D, and TRIDENT-C variants. The results compare
the influence of translation-model selection and classification strategy
on detection performance across languages. \textbf{Bold} values indicate the best performance within each classification strategy, while \textcolor{blue}{blue} values highlight the best performance across RTT models for each classification strategy and dataset.}
\label{tab:multilang_results}

\renewcommand{\arraystretch}{1.15}
\setlength{\tabcolsep}{3pt}
\rowcolors{3}{gray!30}{white}

\resizebox{\textwidth}{!}{%
\begin{tabular}{ll*{14}{c}}
\toprule

\rowcolor{blue!50!black}
\textcolor{white}{\textbf{Model Type}} &
\textcolor{white}{\textbf{Model}} &
\multicolumn{2}{c}{\textcolor{white}{\textbf{Urdu}}} &
\multicolumn{2}{c}{\textcolor{white}{\textbf{Arabic}}} &
\multicolumn{2}{c}{\textcolor{white}{\textbf{Korean}}} &
\multicolumn{2}{c}{\textcolor{white}{\textbf{Russian}}} &
\multicolumn{2}{c}{\textcolor{white}{\textbf{Bulgarian}}} &
\multicolumn{2}{c}{\textcolor{white}{\textbf{Chinese}}} &
\multicolumn{2}{c}{\textcolor{white}{\textbf{RAID (English)}}} \\

\cmidrule(lr){3-4}
\cmidrule(lr){5-6}
\cmidrule(lr){7-8}
\cmidrule(lr){9-10}
\cmidrule(lr){11-12}
\cmidrule(lr){13-14}
\cmidrule(lr){15-16}

\rowcolor{blue!50!black}
& &
\textcolor{white}{\textbf{Acc}} & \textcolor{white}{\textbf{F1}} &
\textcolor{white}{\textbf{Acc}} & \textcolor{white}{\textbf{F1}} &
\textcolor{white}{\textbf{Acc}} & \textcolor{white}{\textbf{F1}} &
\textcolor{white}{\textbf{Acc}} & \textcolor{white}{\textbf{F1}} &
\textcolor{white}{\textbf{Acc}} & \textcolor{white}{\textbf{F1}} &
\textcolor{white}{\textbf{Acc}} & \textcolor{white}{\textbf{F1}} &
\textcolor{white}{\textbf{Acc}} & \textcolor{white}{\textbf{F1}} \\

\midrule

\multirow{4}{*}{Static Threshold}
& GPT-3.5
& \textcolor{blue}{0.90} & \textcolor{blue}{0.88} & \textcolor{blue}{0.91} & \textcolor{blue}{0.85} & \textcolor{blue}{0.89} & 0.81
& \textcolor{blue}{0.77} & \textbf{0.76} & \textcolor{blue}{0.90} & \textcolor{blue}{0.89} & \textcolor{blue}{0.87} & \textcolor{blue}{0.86}
& 0.82 & 0.81 \\

TRIDENT-S & LLaMA 3.3
& 0.86 & 0.83 & 0.88 & 0.81 & \textcolor{blue}{0.89} & \textcolor{blue}{0.84}
& 0.75 & 0.71 & 0.72 & 0.74 & 0.84 & \textcolor{blue}{0.86}
& \textbf{0.86} & \textcolor{blue}{0.87} \\

Static Threshold & Gemma 3
& 0.85 & 0.82 & 0.77 & 0.73 & 0.81 & 0.72
& 0.70 & 0.69 & 0.87 & 0.88 & 0.83 & 0.82
& 0.71 & 0.72 \\

& Phi-4
& 0.84 & 0.81 & 0.79 & 0.79 & 0.82 & 0.78
& 0.71 & 0.68 & 0.81 & 0.81 & 0.81 & 0.79
& 0.76 & 0.75 \\

\midrule

\multirow{4}{*}{Dynamic Threshold}
& GPT-3.5
& \textbf{0.86} & \textbf{0.81} & \textbf{0.87} & \textbf{0.83} & \textbf{0.86} & 0.81
& 0.76 & 0.75 & \textbf{0.88} & \textbf{0.88} & \textbf{0.84} & \textbf{0.83}
& \textbf{0.82} & \textbf{0.79} \\

TRIDENT-D & LLaMA 3.3
Dynamic Threshold & 0.84 & \textbf{0.81} & 0.86 & 0.82 & 0.87 & \textcolor{blue}{0.84}
& 0.75 & 0.76 & 0.81 & 0.80 & 0.81 & 0.81
& 0.78 & \textbf{0.79} \\

& Gemma 3
& 0.81 & 0.79 & 0.76 & 0.73 & 0.81 & 0.75
& \textcolor{blue}{0.77} & \textcolor{blue}{0.77} & 0.85 & 0.84 & 0.82 & 0.81
& 0.79 & \textbf{0.79} \\

& Phi-4
& 0.83 & 0.80 & 0.80 & 0.78 & 0.83 & 0.79
& 0.69 & 0.69 & 0.77 & 0.79 & 0.74 & 0.74
& 0.75 & 0.76 \\

\midrule

\multirow{4}{*}{Logistic Regression}
& GPT-3.5
& \textbf{0.89} & \textbf{0.87} & \textbf{0.88} & \textcolor{blue}{0.85} & 0.86 & 0.79
& 0.76 & \textbf{0.76} & \textbf{0.88} & \textbf{0.87} & \textbf{0.84} & \textbf{0.83}
& \textcolor{blue}{0.90} & \textbf{0.88} \\

TRIDENT-C & LLaMA 3.3
& 0.86 & 0.83 & 0.85 & 0.79 & \textbf{0.87} & \textbf{0.83}
& \textcolor{blue}{0.77} & \textbf{0.76} & 0.85 & 0.85 & 0.83 & 0.82
& 0.86 & 0.86 \\

Logistic Regression & Gemma 3
& 0.82 & 0.79 & 0.75 & 0.71 & 0.79 & 0.70
& 0.73 & 0.75 & 0.81 & 0.84 & \textbf{0.84} & 0.81
& 0.86 & \textbf{0.88} \\

& Phi-4
& 0.83 & 0.81 & 0.79 & 0.78 & 0.82 & 0.77
& 0.69 & 0.70 & 0.77 & 0.78 & 0.75 & 0.76
& 0.81 & 0.83 \\

\bottomrule
\end{tabular}%
}
\end{table*}

\begin{figure}[t]
    \centering
    \includegraphics[width=0.85\textwidth]{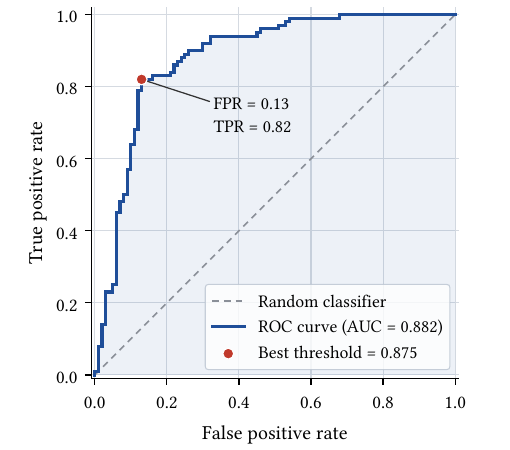} 
    \caption{Receiver operating characteristic (ROC) curve of TRIDENT on the SHAKTI dataset, showing the trade-off between the true positive rate (TPR) and false positive rate (FPR). The model achieves an area under the curve (AUC) of 0.882, with a selected decision threshold of 0.875.}
    \label{tab:threshold}
\end{figure}

\section{Results and Discussion}
Table~\ref{tab:llm_detection_results} presents a comparison between the proposed and baseline models on our proposed dataset, SHAKTI, and the dataset from the literature, AGhi,across various genres. The conventional statistical detection methods, including Log-Likelihood, Rank, Log-Rank, and Entropy, exhibit limited effectiveness in distinguishing human-written from LLM-generated text across different genres and datasets. In comparison, DetectGPT demonstrates improved detection capability by exploiting perturbation-based characteristics of generated text, while RAIDAR achieves further improvements through rewriting-based detection. This progression suggests that methods leveraging transformations of the input text provide more discriminative signals than those relying solely on token-level statistical properties. Nevertheless, the performance of these approaches remains limited, highlighting the challenges of reliably detecting LLM-generated Hindi text using conventional statistical and rewriting-based methods. In contrast to conventional statistical and rewriting-based detection methods, XLM-RoBERTa demonstrates substantially stronger detection performance across different genres and datasets, highlighting the effectiveness of contextual representations for distinguishing human-written from LLM-generated text. Incorporating additional linguistic features generally improves its performance, suggesting that handcrafted linguistic characteristics provide complementary information beyond pretrained contextual embeddings. However, these improvements are not consistent across all genres, indicating that the contribution of linguistic features may vary depending on the characteristics of the underlying text. This variability highlights the challenges of developing robust detection models that generalize consistently across diverse linguistic contexts and genres. 





From Table~\ref{tab:llm_detection_results}, it is evident that the proposed TRIDENT framework consistently outperforms conventional statistical, rewriting-based, and supervised detection baseline models from litrature across all evaluated genres and datasets, demonstrating the effectiveness of RTT-induced semantic variations as discriminative signals for detecting LLM-generated text. Among the variants of TRIDENT, \textit{TRIDENT-S (Static Threshold-Based Classification)} achieves the strongest overall performance, followed by \textit{TRIDENT-C (Logistic Regression with Multi-Distance RTT Features)} and \textit{TRIDENT-D (Dynamic Threshold-Based Classification)}. Interestingly, the superior performance of static thresholding (TRIDENT-S) suggests that RTT-based similarity captures meaningful differences between human-written and LLM-generated text without requiring complex decision boundaries. Although dynamic thresholding (TRIDENT-D) provides a data-dependent alternative to fixed threshold selection, its comparatively lower performance suggests that adaptive threshold estimation does not necessarily improve the performance of TRIDENT. Furthermore, TRIDENT maintains competitive performance across different genres and the independently sourced Hindi dataset, suggesting the potential of RTT-based detection beyond a single dataset. From these observations, we conclude that translation-induced representational changes (TRIDENT) are highly proficient in detecting AI-generated content. The superior performance of TRIDENT stems from several factors: (i) round-trip translation via English leverages the rich semantic and stylistic representations learned by LLMs predominantly trained on English corpora; (ii) the zero-shot formulation enables strong generalisation across domains without task-specific fine-tuning. In contrast, TRIDENT-S (Static Threshold) uses round-trip translation captures deeper semantic shifts and stylistic instability, making it resilient to attacks where simple rewriting fails to fool the system.

\begin{table*}[t]
\centering
\small
\caption{Comparative analysis of the robustness of the three TRIDENT
variants---static threshold (TRIDENT-S), dynamic threshold (TRIDENT-D),
and logistic regression (TRIDENT-C)---under five adversarial attacks on
the SHAKTI dataset. Accuracy (Acc) and Macro-F1 are reported for
rewriting, homoglyph substitution, synonym replacement, misspelling,
and whitespace perturbations.}
\label{tab:detectrtt_adversarial_attacks}

\renewcommand{\arraystretch}{1.15}
\setlength{\tabcolsep}{6pt}
\rowcolors{3}{gray!30}{white}

\begin{tabular}{lcccccc}
\toprule

\rowcolor{blue!50!black}
\textcolor{white}{\textbf{Attack}}
& \multicolumn{2}{c}{\textcolor{white}{\textbf{TRIDENT-S}}}
& \multicolumn{2}{c}{\textcolor{white}{\textbf{TRIDENT-D}}}
& \multicolumn{2}{c}{\textcolor{white}{\textbf{TRIDENT-C}}} \\

\rowcolor{blue!50!black}
& \textcolor{white}{\textbf{Acc}}
& \textcolor{white}{\textbf{Macro-F1}}
& \textcolor{white}{\textbf{Acc}}
& \textcolor{white}{\textbf{Macro-F1}}
& \textcolor{white}{\textbf{Acc}}
& \textcolor{white}{\textbf{Macro-F1}} \\
\midrule

Rewrite
& 0.88 & 0.815
& 0.71 & 0.755
& 0.75 & 0.795 \\

Homoglyph
& 0.80 & 0.745
& 0.73 & 0.675
& 0.77 & 0.715 \\

Synonym Replacement
& 0.72 & 0.665
& 0.66 & 0.605
& 0.70 & 0.645 \\

Misspelling
& 0.52 & 0.475
& 0.45 & 0.405
& 0.49 & 0.445 \\

Whitespace
& 0.76 & 0.705
& 0.70 & 0.645
& 0.74 & 0.685 \\

\bottomrule
\end{tabular}
\end{table*}






\subsection{Sensitivity to Translation Models and Multilingual Adaptability}
To study the sensitivity and impact of the translation model used for round-trip translation and its adaptability to TRIDENT across diverse languages, we conducted an empirical study evaluating TRIDENT's performance by varying the text translation model across multiple languages. Table~\ref{tab:multilang_results} presents the performance of TRIDENT across three language datasets Urdu\cite{ali2025hlu}, Arabic~\cite{al2025arabic}, and Korean~\cite{park2025katfishnet} using four LLM-based translation models: GPT-3.5, LLaMA-3.3, Gemma-3, and Phi-4.

Table~\ref{tab:multilang_results} reveals noticeable variations in detection performance across translation models, indicating that the effectiveness of RTT-based detection is influenced by the underlying translator. GPT-3.5-based translation generally yields the strongest performance, particularly with the static-threshold variant, while LLaMA demonstrates competitive and relatively consistent results across languages. In contrast, Gemma and Phi exhibit comparatively weaker performance in several settings, suggesting that different translation models may introduce distinct degrees of semantic and stylistic variation while round-trip translations. Interestingly, static thresholding generally outperforms the other classification strategies, although dynamic thresholding remains competitive in certain configurations. From such observations, we can conclude that the selection of the translation model plays an important role in determining the discriminative utility of RTT-induced representations.

From Table~\ref{tab:multilang_results}, it is apparent that across Urdu, Arabic, and Korean, the TRIDENT dataset demonstrates promising detection performance, suggesting that translation-induced representational changes provide useful signals beyond Hindi. However, performance varies across languages and translation models, indicating that the effectiveness of RTT-based detection is not entirely language-independent. In particular, certain translation models perform consistently across languages, whereas others exhibit greater variability, potentially reflecting differences in their multilingual translation capabilities and the linguistic characteristics of the evaluated datasets. The results also suggest that the relative effectiveness of the classification variants depends on the language–translator combination, highlighting the importance of evaluating multilingual detection under diverse translation configurations rather than relying on a single translator.

Furthermore, Figure~\ref{fig:mbart} examines the sensitivity of TRIDENT (Static Threshold) to the choice of round-trip translation model, comparing the conventional machine translation model mBART with LLM-based translators, including GPT-3.5, LLaMA-3.3, Gemma-3, and Phi-4, across the four genres of the proposed SHAKTI dataset. The results demonstrate that LLM-based translators consistently outperform mBART, suggesting that the linguistic transformations introduced by different translation architectures have a substantial influence on the discriminative effectiveness of RTT-based detection. Among the evaluated translators, GPT-3.5 exhibits the most balanced performance across genres, whereas Phi-4 achieves particularly strong performance on Wiki and QA but shows greater variability on News and Essay. Similarly, LLaMA-3.3 and Gemma-3 perform more effectively on Wiki than on QA and News, indicating that translation-model effectiveness is influenced by genre-specific linguistic characteristics. These observations are consistent with our hypothesis that differences in RTT-induced representational stability between human-written and LLM-generated texts provide useful detection signals. However, the variation across translators suggests that these signals depend on the nature of the translation transformations rather than being uniformly preserved across models. Complementing this analysis, the lexical divergence results in Table~\ref{tab:information_divergence} (Section~\ref{app:kl}) reveal differences in the similarity between human-written and LLM-generated lexical distributions across generators, providing an additional perspective on their linguistic characteristics. Overall, these findings highlight the importance of translation-model selection and genre-specific evaluation when assessing the effectiveness and generalizability of TRIDENT.

In summary, these findings demonstrate the potential of TRIDENT for multilingual LLM-generated text detection while highlighting its sensitivity to translation model selection. Although the framework remains applicable across the evaluated languages, its effectiveness depends on the interaction among the underlying translation model, the target language, and the classification strategy.

\subsection{Performance Under Different Adversarial Perturbations}

To investigate the robustness of TRIDENT under adversarial conditions, we evaluate its three variants against five perturbation types in the SHAKTI dataset: rewriting, homoglyph substitution, synonym replacement, misspelling, and whitespace perturbation. These experiments examine how different lexical, semantic, and character-level modifications affect the RTT-derived features used to distinguish human-written from LLM-generated text. For each attack type, adversarial examples are constructed by applying the corresponding perturbation to the original test samples while preserving their authorship labels. The models are trained on clean data and evaluated independently on the test sets corresponding to each perturbation, allowing us to assess robustness under distribution shifts introduced by individual attacks without incorporating adversarial examples into training.

As shown in Table~\ref{tab:detectrtt_adversarial_attacks}, TRIDENT-S consistently outperforms TRIDENT-D and TRIDENT-C across all evaluated adversarial attacks in terms of accuracy and class-wise F1 scores. Among these attacks, homoglyph substitution has the least impact on detection performance, whereas misspelling poses the greatest challenge across all three variants. This vulnerability suggests that character-level perturbations, particularly typographical errors, can disrupt the lexical and statistical patterns underlying the detection process. As further illustrated in Figure~\ref{fig:error}, misspellings and typographical errors contribute to misclassification, highlighting the sensitivity of the proposed framework to such perturbations. Rewriting and synonym replacement also adversely affect detection performance, while homoglyph substitution and whitespace perturbations are comparatively less challenging. In summary, TRIDENT-S demonstrates the strongest robustness among the three variants, although the observed performance degradation under certain attacks highlights opportunities to improve adversarial resilience.
\subsection{Cross-LLM Generalization Evaluation}
\label{sec:cross_llm}

To evaluate the generalization capability of TRIDENT-C across unseen text generators, we conduct a cross-LLM evaluation on the SHAKTI dataset using a leave-one-generator-out protocol. In each experiment, one LLM generator is excluded from training and reserved exclusively for testing, while texts from the remaining generators are used to train the logistic regression classifier. GPT-3.5 is employed as the fixed RTT model across all experiments to isolate the effect of variations in the text generators. Table~\ref{tab:cross_llm} presents the TRIDENT performance across the evaluated held-out generators. TRIDENT-C demonstrates promising cross-generator generalization, maintaining consistent performance on AI-generated texts from previously unseen generators. The strongest performance is observed on LLaMA-3.2-generated texts, whereas GPT-3.5-generated texts pose the greatest challenge. These findings suggest that RTT-derived features capture transferable detection signals beyond generator-specific characteristics, although performance variations indicate some sensitivity to the underlying text generator. Nevertheless, the variation in performance across held-out generators indicates that detection effectiveness remains influenced by generator-specific characteristics. In summary, the results provide evidence that TRIDENT-C can generalise to previously unseen text generators without requiring generator-specific retraining.

\begin{table*}[t]
\centering
\small
\caption{Comparative analysis of cross-LLM generalization performance
of TRIDENT-C on the SHAKTI dataset. Each experiment excludes
one LLM generator from training and evaluates detection performance
on the unseen generator. GPT-3.5 is used as the fixed round-trip
translation (RTT) model across all experiments. Accuracy, human-class
F1 (F1-H), and AI-generated-class F1 (F1-AI) are reported to assess
the model's ability to generalize across different text generators.}
\label{tab:cross_llm}

\renewcommand{\arraystretch}{1.15}
\setlength{\tabcolsep}{5pt}
\rowcolors{3}{gray!30}{white}

\begin{tabular}{llcccc}
\toprule

\rowcolor{blue!50!black}
\textcolor{white}{\textbf{RTT Model}} &
\textcolor{white}{\textbf{Training Generators}} &
\textcolor{white}{\textbf{Test Generator}} &
\textcolor{white}{\textbf{Accuracy}} &
\textcolor{white}{\textbf{F1-H}} &
\textcolor{white}{\textbf{F1-AI}} \\
\midrule

GPT-3.5 &
GPT-3.5, Mistral, Gemma-3 &
LLaMA-3.2 & 0.87 & 0.83 & 0.89 \\

GPT-3.5 &
GPT-3.5, LLaMA-3.2, Gemma-3 &
Mistral & 0.84 & 0.80 & 0.89 \\

GPT-3.5 &
GPT-3.5, LLaMA-3.2, Mistral &
Gemma-3 & 0.86 & 0.82 & 0.88 \\

GPT-3.5 &
LLaMA-3.2, Mistral, Gemma-3 &
GPT-3.5 & 0.82 & 0.78 & 0.87 \\

\bottomrule
\end{tabular}
\end{table*}

\begin{figure}[t]
    \centering
    \includegraphics[width=0.95\textwidth]{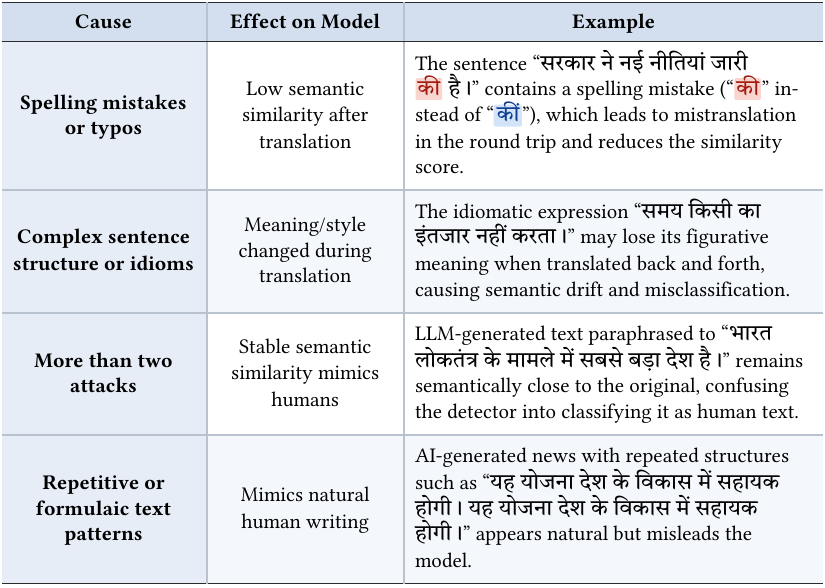} 
   \caption{\small Representative misclassification cases produced by TRIDENT on the SHAKTI dataset. The examples illustrate how repetitive or formulaic LLM-generated text and linguistic or translation-related variations can lead to incorrect authorship predictions.}
    \label{fig:error} 
\end{figure}

\subsection{Error Analysis}
\label{sec:error_ana}
To better understand the limitations of TRIDENT-S, we conduct a qualitative error analysis of samples misclassified by TRIDENT, examining how linguistic complexity, translation-induced artifacts, and adversarial perturbations contribute to incorrect predictions, as illustrated in Figure~\ref{fig:error}.

Figure~\ref{fig:error} presents examples of misclassifications by TRIDENT-S, where linguistic variations and translation artifacts lead to incorrect classification by TRIDENT, and attacks might also be misclassified. Our error analysis reveals that misclassifications frequently arise from linguistic irregularities and translation-induced variations. In particular, human-written texts containing spelling errors or typographical inconsistencies tend to undergo greater changes during round-trip translation, potentially altering their semantic representations and leading to incorrect classification as LLM-generated text. Similarly, informal expressions, dialectal variations, and region-specific phrases introduce additional ambiguity during translation, making it difficult to preserve the original meaning. These observations suggest that TRIDENT-S is sensitive to linguistic phenomena that are not consistently preserved across translation stages, highlighting a potential limitation when processing informal or linguistically diverse human-written content. 
Further analysis reveals that texts containing complex grammatical constructions, idiomatic expressions, and multiple negations are particularly susceptible to translation-induced semantic shifts, contributing to both false-positive and false-negative predictions. Interestingly, heavily paraphrased LLM-generated texts may retain their underlying semantic representations despite substantial surface-level modifications, making them difficult to distinguish from human-written content. Similarly, repetitive or formulaic structures in LLM-generated texts can exhibit linguistic characteristics resembling those of human-written texts, further reducing their discriminability. These observations highlight the limitations of relying solely on RTT-induced semantic variations, particularly when linguistic complexity and adversarial paraphrasing affect the relationship between semantic stability and text authorship.

\section{Conclusion and Future Work}
In this study, we introduce \textsc{SHAKTI}, a large-scale Hindi benchmark for LLM-generated text detection, comprising human-written content across diverse genres and corresponding LLM-generated texts from different LLM families. Additionally, the \textsc{SHAKTI} dataset is curated using diverse prompting strategies and adversarial transformations, enabling systematic evaluation of detection methods and fostering the development of models for AI-generated content detection in both conventional and adversarial settings. Next, we conduct a comprehensive linguistic analysis of human-written and LLM-generated Hindi texts across structural, lexical, readability, and language-specific features. Our findings reveal both similarities and differences in their linguistic characteristics, with certain LLMs producing texts that closely resemble human-written content in terms of lexical distributions. These observations highlight the limitations of relying solely on surface-level linguistic features for reliable detection and motivate the investigation of complementary detection signals.

Furthermore, this study proposes \textsc{TRIDENT}, a round-trip translation-based framework and its variants for detecting LLM-generated text. Extensive experiments on SHAKTI, an existing Hindi benchmark, and datasets in Urdu, Arabic, and Korean demonstrated the effectiveness and multilingual applicability of the proposed approach.  \textsc{TRIDENT} consistently outperformed conventional statistical, rewriting-based, and supervised detection baselines across the evaluated Hindi datasets. Among its three variants, \textsc{TRIDENT-S} (Static Threshold-Based Classification), \textsc{TRIDENT-D} (Dynamic Threshold-Based Classification), and \textsc{TRIDENT-C} (Logistic Regression-Based Classification with Multi-Distance RTT Features), static thresholding (\textsc{TRIDENT-S}) achieved the strongest overall performance, suggesting that RTT-induced semantic similarity provides a discriminative signal without requiring complex classification boundaries.

Further analysis revealed that detection performance depends on the choice of translation model, with LLM-based translators generally outperforming the conventional machine translation model mBART. The results also highlighted genre-dependent variations and performance differences across languages, emphasising the importance of selecting a translation model when applying RTT-based detection in multilingual settings. Nevertheless, qualitative error analysis identified challenges involving linguistic complexity, translation artefacts, and adversarial paraphrasing. Overall, our findings establish round-trip translation-induced representational changes as a promising signal for LLM-generated text detection and motivate future research toward more efficient, linguistically adaptive, and robust multilingual detection frameworks.

\section*{Limitations}

While SHAKTI and TRIDENT advance LLM-generated text detection for low-resource languages, several limitations remain. First, the dataset and primary analysis focus on Hindi, and although additional languages are evaluated, out-of-domain and cross-lingual generalization is limited. Detection performance may also depend on the LLMs and domains represented in the data, reducing robustness to unseen domains and future models. Second, large language models exhibit behavior drift over time. Although TRIDENT mitigates reliance on model-specific artifacts through round-trip translation and prompt-agnostic similarity modeling, it does not fully eliminate the impact of substantial shifts in LLM generation behavior. Third, our similarity modeling primarily relies on cosine similarity between sentence embeddings, leaving room for alternative semantic comparison measures. Finally, human-written texts were not generated using the same prompts as LLM-generated content. While topical and stylistic alignment was maintained, this mismatch may introduce distributional differences unrelated to authorship.

 \section*{Ethics and Privacy Statement}
The \textit{SHAKTI} is curated without collecting any personal information for LLM-generated text detection in the Hindi language.
\textit{SHAKTI} is curated based on publicly available news article corpus~\cite{kumar2024fake}, Wikipedia articles\cite{wikipediaHindiHome}, Question answering, essays and LLM-generated content produced via controlled prompting of large language models. The dataset is released under the Creative Commons Attribution 4.0 International license, and the accompanying code is released under the MIT license.

\section*{Declaration of Generative AI Use}
Generative AI tools were used to assist with language refinement, grammatical corrections, manuscript readability, and the preparation and formatting of figures and tables. All AI-assisted content was reviewed and verified by the authors, who retain full responsibility for the accuracy, originality, scientific integrity, experimental results, interpretations, and conclusions presented in the manuscript.

\bibliographystyle{unsrt}  
\bibliography{references}

\section*{Appendix}

\section{Prompt used to generate Dataset}
\label{Appendix:propmt}
\begin{tcolorbox}[
    enhanced,
    breakable,
    colback=gray!3,
    colframe=black!55,
    coltitle=black,
    colbacktitle=gray!15,
    title={Wikipedia Generation Prompt},
    fonttitle=\bfseries\small,
    boxrule=0.5pt,
    arc=2pt,
    left=8pt,
    right=8pt,
    top=6pt,
    bottom=6pt
]
\small

\textbf{Role:}
You are a Hindi encyclopedic content writer.

\vspace{0.15cm}
\textbf{Task:}
Generate an informative Hindi passage about the specified
topic in an encyclopedic style.

\vspace{0.15cm}
\textbf{Input:}

Title: \texttt{[TITLE]}

\vspace{0.15cm}
\textbf{Instructions:}
\begin{itemize}[
    leftmargin=*,
    nosep,
    topsep=2pt
]
    \item Use a neutral, factual, and informative tone.
    \item Organize the content logically.
    \item Include relevant background, key facts, and
          explanations.
    \item Use formal and grammatically correct Hindi.
    \item Avoid personal opinions, conversational expressions,
          and repetitive statements.
    \item Do not invent specific facts or references.
\end{itemize}

\vspace{0.15cm}
\textbf{Output:}
Return only the generated Hindi passage.

\end{tcolorbox}

\begin{tcolorbox}[
    enhanced,
    breakable,
    colback=gray!3,
    colframe=black!55,
    coltitle=black,
    colbacktitle=gray!15,
    title={Question Answering (QA) Prompt},
    fonttitle=\bfseries\small,
    boxrule=0.5pt,
    arc=2pt,
    left=8pt,
    right=8pt,
    top=6pt,
    bottom=6pt
]
\small

\textbf{Role:}
You are a Hindi question-answering assistant specializing in Indian history.

\vspace{0.15cm}
\textbf{Task:}
Answer the given question in Hindi.

\vspace{0.15cm}
\textbf{Input:}

Question: \texttt{[QUESTION]}

\vspace{0.15cm}
\textbf{Requirements:}
\begin{itemize}[
    leftmargin=*,
    nosep,
    topsep=2pt
]
    \item Provide a direct, relevant, and informative answer.
    \item Include appropriate historical facts and explanations.
    \item Use natural and grammatically correct Hindi.
    \item Keep the answer concise while including essential information.
    \item Avoid unrelated details and unsupported claims.
    \item Do not repeat the question in the answer.
\end{itemize}

\vspace{0.15cm}
\textbf{Output:}
Return only the Hindi answer.

\end{tcolorbox}

\begin{tcolorbox}[
    enhanced,
    breakable,
    colback=gray!3,
    colframe=black!55,
    coltitle=black,
    colbacktitle=gray!15,
    title={News Generation Prompt},
    fonttitle=\bfseries\small,
    boxrule=0.5pt,
    arc=2pt,
    left=8pt,
    right=8pt,
    top=6pt,
    bottom=6pt
]
\small

\textbf{Role:}
You are a Hindi news article writer.

\vspace{0.15cm}
\textbf{Task:}
Generate a Hindi news article based on the provided topic or source information.

\vspace{0.15cm}
\textbf{Input:}

Topic or source text: \texttt{[HEADLINES]}

\vspace{0.15cm}
\textbf{Requirements:}
\begin{itemize}[
    leftmargin=*,
    nosep,
    topsep=2pt
]
    \item Use a formal, journalistic writing style.
    \item Present information coherently and objectively.
    \item Include relevant details and maintain consistency throughout the article.
    \item Use natural and grammatically correct Hindi.
    \item Avoid unsupported claims and unnecessary repetition.
    \item Do not copy the source text verbatim.
\end{itemize}

\vspace{0.15cm}
\textbf{Output:}
Return only the generated Hindi news article.

\end{tcolorbox}

\begin{tcolorbox}[
    enhanced,
    breakable,
    colback=gray!3,
    colframe=black!55,
    coltitle=black,
    colbacktitle=gray!15,
    title={Essay Generation Prompt},
    fonttitle=\bfseries\small,
    boxrule=0.5pt,
    arc=2pt,
    left=8pt,
    right=8pt,
    top=6pt,
    bottom=6pt
]
\small

\textbf{Role:}
You are a Hindi essay writer.

\vspace{0.15cm}
\textbf{Task:}
Write a coherent Hindi essay on the given topic.

\vspace{0.15cm}
\textbf{Input:}

Essay topic: \texttt{[TOPIC]}

\vspace{0.15cm}
\textbf{Requirements:}
\begin{itemize}[
    leftmargin=*,
    nosep,
    topsep=2pt
]
    \item Introduce the topic and develop the discussion in a logical sequence.
    \item Present relevant arguments, examples, and explanations.
    \item Use natural Hindi with varied vocabulary and sentence structures.
    \item Maintain coherence and consistency throughout the essay.
    \item Include a concluding paragraph summarizing the main ideas.
    \item Avoid unnecessary repetition and unrelated content.
\end{itemize}

\vspace{0.15cm}
\textbf{Output:}
Return only the complete Hindi essay.

\end{tcolorbox}

\textbf{Adversarial attacks prompt}

\begin{tcolorbox}[
    enhanced,
    breakable,
    colback=gray!3,
    colframe=black!55,
    coltitle=black,
    colbacktitle=gray!15,
    title={Rewrite Attack Prompt},
    fonttitle=\bfseries\small,
    boxrule=0.5pt,
    arc=2pt,
    left=8pt,
    right=8pt,
    top=6pt,
    bottom=6pt
]
\small

\textbf{Role:}
You are a Hindi text transformation assistant.

\vspace{0.15cm}
\textbf{Task:}
Rewrite the given Hindi text while preserving its original meaning.

\vspace{0.15cm}
\textbf{Input:}

Original Text: \texttt{[ORIGINAL TEXT]}

\vspace{0.15cm}
\textbf{Instructions:}
\begin{itemize}[
    leftmargin=*,
    nosep,
    topsep=2pt
]
    \item Express the same information using different words and sentence structures.
    \item Preserve the original meaning, facts, entities, and intent.
    \item Maintain fluent and grammatically correct Hindi.
    \item Do not add new information or remove important details.
    \item Avoid copying the original wording wherever possible.
    \item Do not translate the text into another language.
    \item Keep the output approximately the same length as the input.
\end{itemize}

\vspace{0.15cm}
\textbf{Output:}
Return only the rewritten Hindi text.

\end{tcolorbox}

\begin{tcolorbox}[
    enhanced,
    breakable,
    colback=gray!3,
    colframe=black!55,
    coltitle=black,
    colbacktitle=gray!15,
    title={Homoglyph Substitution Attack Prompt},
    fonttitle=\bfseries\small,
    boxrule=0.5pt,
    arc=2pt,
    left=8pt,
    right=8pt,
    top=6pt,
    bottom=6pt
]
\small

\textbf{Role:}
You are a character-level Hindi text transformation assistant.

\vspace{0.15cm}
\textbf{Task:}
Apply homoglyph substitutions to the given Hindi text.

\vspace{0.15cm}
\textbf{Input:}

Original Text: \texttt{[ORIGINAL TEXT]}

\vspace{0.15cm}
\textbf{Instructions:}
\begin{itemize}[
    leftmargin=*,
    nosep,
    topsep=2pt
]
    \item Replace eligible characters with visually similar characters from the specified mapping.
    \item Preserve the original word order, spacing, punctuation, and sentence structure.
    \item Do not rewrite, paraphrase, translate, or correct the text.
    \item Do not substitute characters outside the permitted mapping.
    \item If no eligible character is present, return the original text unchanged.
\end{itemize}

\vspace{0.15cm}
\textbf{Output:}
Return only the transformed text.

\end{tcolorbox}

\begin{tcolorbox}[
    enhanced,
    breakable,
    colback=gray!3,
    colframe=black!55,
    coltitle=black,
    colbacktitle=gray!15,
    title={Synonym Replacement Attack Prompt},
    fonttitle=\bfseries\small,
    boxrule=0.5pt,
    arc=2pt,
    left=8pt,
    right=8pt,
    top=6pt,
    bottom=6pt
]
\small

\textbf{Role:}
You are a Hindi lexical substitution assistant.

\vspace{0.15cm}
\textbf{Task:}
Replace selected words in the given Hindi text with appropriate synonyms while preserving the original meaning.

\vspace{0.15cm}
\textbf{Input:}

Original Text: \texttt{[ORIGINAL TEXT]}

\vspace{0.15cm}
\textbf{Instructions:}
\begin{itemize}[
    leftmargin=*,
    nosep,
    topsep=2pt
]
    \item Replace the specified proportion of eligible words with suitable Hindi synonyms.
    \item Preserve the meaning and grammatical correctness of each sentence.
    \item Maintain the original sentence order, named entities, and factual content.
    \item Do not replace words that lack suitable synonyms.
    \item Do not restructure sentences or rewrite the entire passage.
    \item Avoid introducing ambiguity or changing the intended meaning.
\end{itemize}

\vspace{0.15cm}
\textbf{Output:}
Return only the transformed Hindi text.

\end{tcolorbox}

\begin{tcolorbox}[
    enhanced,
    breakable,
    colback=gray!3,
    colframe=black!55,
    coltitle=black,
    colbacktitle=gray!15,
    title={Misspelling Attack Prompt},
    fonttitle=\bfseries\small,
    boxrule=0.5pt,
    arc=2pt,
    left=8pt,
    right=8pt,
    top=6pt,
    bottom=6pt
]
\small

\textbf{Role:}
You are a Hindi text perturbation assistant.

\vspace{0.15cm}
\textbf{Task:}
Introduce controlled spelling errors into the given Hindi text.

\vspace{0.15cm}
\textbf{Input:}

Original Text: \texttt{[ORIGINAL TEXT]}

\vspace{0.15cm}
\textbf{Instructions:}
\begin{itemize}[
    leftmargin=*,
    nosep,
    topsep=2pt
]
    \item Introduce spelling errors in the specified proportion of eligible words.
    \item Apply only the specified error operations, such as character deletion, insertion, or substitution.
    \item Preserve the original meaning as much as possible.
    \item Do not intentionally change named entities, numbers, or punctuation unless specified.
    \item Do not paraphrase, translate, or reorder sentences.
    \item Leave unaffected words unchanged.
\end{itemize}

\vspace{0.15cm}
\textbf{Output:}
Return only the perturbed Hindi text.

\end{tcolorbox}

\begin{tcolorbox}[
    enhanced,
    breakable,
    colback=gray!3,
    colframe=black!55,
    coltitle=black,
    colbacktitle=gray!15,
    title={Whitespace Manipulation Attack Prompt},
    fonttitle=\bfseries\small,
    boxrule=0.5pt,
    arc=2pt,
    left=8pt,
    right=8pt,
    top=6pt,
    bottom=6pt
]
\small

\textbf{Role:}
You are a character-level text perturbation assistant.

\vspace{0.15cm}
\textbf{Task:}
Modify whitespace in the given Hindi text without changing its words.

\vspace{0.15cm}
\textbf{Input:}

Original Text: \texttt{[ORIGINAL TEXT]}

\vspace{0.15cm}
\textbf{Instructions:}
\begin{itemize}[
    leftmargin=*,
    nosep,
    topsep=2pt
]
    \item Apply the specified whitespace operation, such as inserting extra spaces or modifying spacing at eligible positions.
    \item Use the specified perturbation rate.
    \item Preserve all non-whitespace characters and their order.
    \item Do not insert, delete, or replace letters, diacritics, or punctuation.
    \item Do not paraphrase, translate, or correct the text.
    \item If no eligible position exists, return the original text unchanged.
\end{itemize}

\vspace{0.15cm}
\textbf{Output:}
Return only the whitespace-transformed text.

\end{tcolorbox}

\section{Mathematical Formulation and Estimation of Linguistic Features}

\subsection{Computation of Structural Linguistic Features}
\label{Appenix:structual}
Let $\mathcal{D}_m = \{x_1,x_2,\ldots,x_{N_m}\}$
denote the collection of texts generated by source $m$, where
$m \in \{\text{Human}, \text{Gemma}, \text{GPT}, \text{LLaMA},\text{Mistral}\,\text{ChatGPT}\,\text{Gemini}\,\text{Aya}\,\text{Sarvam}\}$ and $N_m$ is the number of texts in that collection.
For each text $x_i$, we extract the following features.

\paragraph{Character Count.}
The character count measures the total number of characters in a text sample, which may consist of one or more sentences.
Let $x_i = (c_1,c_2,\ldots,c_{L_i})$ denote the character sequence of the \(i\)-th text sample, where \(L_i\) is the total number of characters in the sample. The character count is defined as
\begin{equation}
    C(x_i) = |x_i| = L_i,
    \label{eq:character_count}
\end{equation}
where whitespace and punctuation marks are included in the count. This convention is applied consistently across all text collections.

\paragraph{Word Count.}
The word count measures the total number of word tokens in a text sample. Let $W(x_i) = (w_1,w_2,\ldots,w_{K_i})$ denote the sequence of word tokens obtained by applying a consistent tokenization procedure to the $i$-th text sample $x_i$. The word count is defined as
\begin{equation}
N_{\mathrm{word}}(x_i) = |W(x_i)| = K_i,
\label{eq}
\end{equation}
where $K_i$ denotes the total number of word tokens across all sentences and paragraphs in the text sample.
Word tokens are identified using whitespace-based segmentation, with the same tokenization procedure applied consistently across all text collections.
\paragraph{Sentence Count.}
The sentence count measures the total number of sentences in a text sample. Let $S(x_i)=(s_1,s_2,\ldots,s_{M_i})$ denote the sequence of sentence segments obtained by applying Hindi-specific sentence-boundary detection rules to the $i$-th text sample $x_i$. These rules identify sentence boundaries using punctuation marks such as the Hindi danda, question mark, and exclamation mark. The sentence count is defined as
\begin{equation}
    N_{\mathrm{sent}}(x_i)=|S(x_i)|=M_i,
    \label{eq:sentence_count}
\end{equation}
where $M_i$ denotes the total number of sentences across the entire text sample. The same sentence-segmentation rules are applied consistently across all text collections.

\paragraph{Paragraph Count.}
The paragraph count measures the total number of non-empty
paragraphs in a text sample. Let
$P(x_i) = (p_1,p_2,\ldots,p_{R_i})$
denote the sequence of paragraphs obtained by splitting
the $i$-th text sample $x_i$ at paragraph boundaries
and excluding empty segments. The paragraph count is
defined as
\begin{equation}
    N_{\mathrm{para}}(x_i) = |P(x_i)| = R_i,
    \label{eq:paragraph_count}
\end{equation}
where $R_i$ denotes the total number of non-empty paragraphs in the text sample. The same paragraph segmentation rules are applied consistently across all text collections. Thus, each newline character is treated as a paragraph boundary, and the count is one greater than the number of newline characters.

\paragraph{Special Character Count.}
The special character count measures the number of
non-alphanumeric symbols, excluding whitespace and
punctuation, in a text sample. Let $\mathcal{A}$
denote the set of Unicode letters, combining marks,
digits, and whitespace characters, and let
$\mathcal{Q}$ denote the set of recognized punctuation
symbols. The special character count is defined as
\begin{equation}
    N_{\mathrm{special}}(x_i)
    = \sum_{j=1}^{L_i}
    \mathbb{I}
    \left(c_j \notin \mathcal{A}
    \cup \mathcal{Q}\right),
    \label{eq:special_character_count}
\end{equation}
where $\mathbb{I}(\cdot)$ is the indicator function,
which equals $1$ when the condition is satisfied
and $0$ otherwise. This definition excludes
Devanagari letters and combining marks, whitespace,
digits, and punctuation from the special character
count.

\paragraph{Punctuation Count.}
The punctuation count measures the total number of punctuation symbols in a text sample. Let
$\mathcal{Q}$ denote the set of Unicode punctuation
characters, including Hindi-specific punctuation
such as the danda (|), as well as commonly used
punctuation marks such as periods, commas, question
marks, and exclamation marks. The punctuation count
is defined as
\begin{equation}
    N_{\mathrm{punct}}(x_i)
    = \sum_{j=1}^{L_i}
    \mathbb{I}(c_j \in \mathcal{Q}),
    \label{eq:punctuation_count}
\end{equation}
where $\mathbb{I}(\cdot)$ denotes the indicator
function. The same Unicode character-classification
rules are applied consistently across all text
collections.

\paragraph{Average Feature Values.}
After extracting the structural features from individual
text samples, we compute the arithmetic mean of each
feature separately for every source $m$. For a given
structural feature $f$, the source-specific average
is defined as
\begin{equation}
    \overline{f}_m =
    \frac{1}{N_m}
    \sum_{i=1}^{N_m} f(x_i),
    \label{eq:average_structural_features}
\end{equation}
where $N_m$ denotes the total number of text samples
originating from source $m$, and $f(x_i)$ represents
the corresponding feature value extracted from the
$i$-th text sample. The resulting average values
are reported in Table~\ref{tab:structural_features},
enabling a comparative analysis of structural
characteristics across human-written and
LLM-generated Hindi texts.

\subsection{Quantifying Lexical Diversity in Human-Written and LLM-Generated Hindi Text}
\label{Appenix:lexical}

Let $x_i$ denote the $i$-th text, and
let $W_i = (w_1,w_2,\ldots,w_{n_i})$ denote its sequence of word
tokens, where $n_i$ is the total number of tokens. The following
features are extracted from each text.

\paragraph{Number of Unique Words.}
The number of unique words measures the vocabulary
size of an individual text sample. Let
$W(x_i) = (w_1,w_2,\ldots,w_{n_i})$
denote the sequence of word tokens in the $i$-th
text sample $x_i$, where $n_i$ represents the total
number of word tokens. Let $V_i$ denote the set
of distinct word types occurring in $x_i$.
The number of unique words is defined as
\begin{equation}
    U(x_i) = |V_i|
    = \left|\{w_j : 1 \leq j \leq n_i\}\right|,
    \label{eq:unique_words}
\end{equation}
where $|V_i|$ denotes the cardinality of the
vocabulary set. A higher value indicates a
greater number of distinct word types, although
this measure is sensitive to the overall text length.

\paragraph{Type-Token Ratio (TTR).}
The type-token ratio measures lexical diversity as the
proportion of distinct word types relative to the total
number of word tokens in a text sample. Let $U(x_i)$
denote the number of unique word types and $n_i$
denote the total number of word tokens in the
$i$-th text sample. TTR is defined as
\begin{equation}
    \mathrm{TTR}(x_i) =
    \frac{U(x_i)}{n_i},
    \label{eq:ttr}
\end{equation}
where $n_i > 0$. TTR ranges from greater than $0$
to $1$, with higher values indicating a greater
proportion of distinct word types. However, TTR
is sensitive to text length, which should be
considered when comparing samples of substantially
different lengths.

\paragraph{Average Word Length.}
Average word length measures the mean number of
characters per word token in a text sample. Let
$W(x_i)=(w_1,w_2,\ldots,w_{n_i})$ denote the
sequence of word tokens in the $i$-th text sample,
and let $\ell(w_j)$ denote the number of characters
in word token $w_j$. The average word length
is defined as
\begin{equation}
    L_{\mathrm{avg}}(x_i) =
    \frac{1}{n_i}
    \sum_{j=1}^{n_i}\ell(w_j),
    \label{eq:avg_word_length}
\end{equation}
where $n_i > 0$. Word length is calculated using
the same character-counting convention across all
text collections. This feature captures variations
in word-form length but does not independently
establish lexical complexity or formality.

\paragraph{Vowel Ratio.}
The vowel ratio measures the relative frequency
of vowel characters in a text sample. Let
$\mathcal{V}$ denote the predefined set of Hindi
vowel characters, including independent Devanagari
vowels and dependent vowel signs (matras). Let
$N_{\mathrm{vowel}}(x_i)$ denote the number of
characters belonging to $\mathcal{V}$, and let
$L_i$ denote the total number of Unicode code
points in the $i$-th text sample. The vowel ratio
is defined as
\begin{equation}
    R_{\mathrm{vowel}}(x_i) =
    \frac{N_{\mathrm{vowel}}(x_i)}{L_i},
    \label{eq:vowel_ratio}
\end{equation}
where $L_i > 0$. The same vowel inventory and
Unicode character-counting convention are applied
consistently across all text collections.
The denominator includes whitespace and punctuation
characters when present.

\paragraph{Punctuation Variety.}
Punctuation variety measures the diversity and
distribution of punctuation symbols in a text
sample. Let $\mathcal{Q}$ denote the predefined
set of punctuation symbols, including Hindi-specific
and standard punctuation marks. Let $K_i$ denote
the number of distinct punctuation types observed
in the $i$-th text sample, and let $p_k$ denote
the relative frequency of punctuation type $k$
among all punctuation occurrences. Punctuation
variety is quantified using Shannon entropy:
\begin{equation}
    H_{\mathrm{punct}}(x_i) =
    -\sum_{k=1}^{K_i} p_k \log p_k,
    \label{eq:punctuation_entropy}
\end{equation}
where the natural logarithm is used. For samples
containing no punctuation, the entropy is defined
as zero. Higher entropy indicates a more diverse
and evenly distributed use of punctuation types,
whereas lower entropy indicates that punctuation
usage is concentrated in fewer types.

\paragraph{Repetition Score.}
The repetition score measures the proportion of
word-token occurrences belonging to word types
that appear more than once in a text sample.
Let $f_i(w)$ denote the frequency of word type
$w$ in the $i$-th text sample, and let $V_i$
denote the set of distinct word types. The set
of repeated word types is defined as
$V_i^{\mathrm{rep}} =
\{w \in V_i : f_i(w) > 1\}$.
The repetition score is calculated as
\begin{equation}
    R_{\mathrm{rep}}(x_i) =
    \frac{
    \sum_{w \in V_i^{\mathrm{rep}}} f_i(w)
    }{n_i},
    \label{eq:repetition_score}
\end{equation}
where $n_i > 0$ denotes the total number of
word tokens. The score ranges from $0$ to $1$,
with higher values indicating a greater proportion
of tokens associated with repeated word types.
Unlike measures that count only duplicate
occurrences beyond the first occurrence, this
formulation includes all occurrences of each
repeated word type.

\subsection{Computation of Readability and Sentence Complexity Measures}
\label{Appenix:Readability}

Let $x_i$ denote the $i$-th text sample containing
$N_i$ word tokens and $S_i$ sentences, where
$N_i > 0$ and $S_i > 0$. Let $w_{ij}$ denote
the $j$-th word token in $x_i$, and let
$\sigma(w_{ij})$ denote its estimated number
of syllables, obtained using a consistent
Hindi-specific syllable-counting procedure.

\paragraph{Average Words per Sentence.}
Average words per sentence measures the mean number
of word tokens across sentences in a text sample.
It is defined as
\begin{equation}
    L_{\mathrm{sent}}(x_i) =
    \frac{N_i}{S_i},
    \label{eq:avg_words_sentence}
\end{equation}
where $N_i$ and $S_i$ represent the total numbers
of word tokens and sentences, respectively.
Higher values indicate longer sentences on average,
although sentence length alone does not establish
syntactic or semantic complexity.

\paragraph{Average Syllables per Word.}
Average syllables per word measures the mean number
of syllables across all word tokens in a text sample.
It is defined as
\begin{equation}
    \overline{\sigma}(x_i) =
    \frac{1}{N_i}
    \sum_{j=1}^{N_i}\sigma(w_{ij}),
    \label{eq:avg_syllables_word}
\end{equation}
where $\sigma(w_{ij})$ denotes the estimated syllable
count of word token $w_{ij}$. For Hindi text,
syllable counts are estimated using a consistent
language-appropriate procedure that accounts for
the characteristics of Devanagari orthography.

\paragraph{Flesch Reading Ease.}
Flesch Reading Ease (FRE) estimates textual readability using average
sentence length and average syllables per word. The conventional
English-language formula is defined as
\begin{equation}
\begin{aligned}
    \mathrm{FRE}(x_i) ={}&
    206.835
    - 1.015\left(\frac{N_i}{S_i}\right)
    - 84.6\left(
    \frac{\sum_{j=1}^{N_i}\sigma(w_{ij})}{N_i}
    \right),
\end{aligned}
\label{eq:flesch_reading_ease}
\end{equation}
where $N_i$ denotes the number of words in text $x_i$, $S_i$ denotes
the number of sentences, $w_{ij}$ denotes the $j$-th word, and
$\sigma(w_{ij})$ represents the estimated number of syllables in that
word. Higher FRE scores conventionally indicate easier-to-read English
text. However, the original formula relies on English-specific
linguistic characteristics and is not directly calibrated for Hindi.
Prior work on readability measures for Bangla and Hindi texts
\cite{sinha2012new} motivates the need for language-specific readability
assessment. Accordingly, we use FRE as an exploratory readability
proxy for Hindi rather than interpreting its scores as a validated
measure of Hindi reading difficulty.

\paragraph{Gunning Fog Index.}
The Gunning Fog Index estimates textual difficulty
based on average sentence length and the proportion
of complex words. Let $C_i$ denote the number of
complex words in text sample $x_i$, identified
according to a predefined syllable-based criterion.
The index is calculated as
\begin{equation}
    \mathrm{Fog}(x_i) =
    0.4\left[
    \frac{N_i}{S_i}
    +100\left(\frac{C_i}{N_i}\right)
    \right].
    \label{eq:gunning_fog}
\end{equation}
Higher values conventionally indicate greater
reading difficulty in English. For Hindi text,
the resulting scores are interpreted as
comparative indicators rather than validated
estimates of reading-grade levels.

\paragraph{SMOG Index.}
The Simple Measure of Gobbledygook (SMOG) Index
estimates readability using the frequency of
polysyllabic words. Let $P_i$ denote the number
of word tokens containing at least three syllables
in text sample $x_i$, according to the adopted
syllable-counting procedure. The SMOG Index
is defined as
\begin{equation}
    \mathrm{SMOG}(x_i) =
    1.0430\sqrt{
    P_i\left(\frac{30}{S_i}\right)
    }+3.1291,
    \label{eq:smog}
\end{equation}
where $S_i$ represents the total number of sentences.
As the original SMOG formula was developed for
English and is conventionally applied to longer
texts, its application to Hindi is considered
exploratory, particularly for shorter samples.

\paragraph{Sentence Length Variance.}
Sentence length variance measures the dispersion
of word counts across sentences within a text
sample. Let $n_{ik}$ denote the number of word
tokens in the $k$-th sentence of text $x_i$,
and let $\overline{n}_i=N_i/S_i$ denote
the average sentence length. The population
variance is defined as
\begin{equation}
    V_{\mathrm{sent}}(x_i) =
    \frac{1}{S_i}
    \sum_{k=1}^{S_i}
    \left(n_{ik}-\overline{n}_i\right)^2,
    \label{eq:sentence_length_variance}
\end{equation}
where $S_i > 0$. Higher values indicate greater
variability in sentence lengths, whereas lower
values indicate more uniform sentence structures.
This feature characterizes sentence-length
distribution but does not independently
establish textual coherence or naturalness.

\subsection{Computation of Indic Script and Semantic Distance Measures}
\label{Appenix:indic}
Let $x_i$ denote the $i$-th text sample containing
$N_i$ word tokens, represented by the token sequence
$W_i=(w_{i1},w_{i2},\ldots,w_{iN_i})$.

\paragraph{Stopword Count.}
Stopwords are frequently occurring function words,
such as postpositions, pronouns, and conjunctions,
that typically contribute less domain-specific
semantic information than content words. Let
$\mathcal{S}$ denote a predefined Hindi stopword
lexicon. The stopword count is defined as
\begin{equation}
    C_{\mathrm{stop}}(x_i) =
    \sum_{j=1}^{N_i}
    \mathbb{I}(w_{ij}\in\mathcal{S}),
    \label{eq:stopword_count}
\end{equation}
where $\mathbb{I}(\cdot)$ is the indicator function,
equal to one when the condition is satisfied and
zero otherwise. A consistent stopword lexicon and
tokenization procedure are applied across all
text collections.

\paragraph{Stopword Density.}
Stopword density measures the proportion of word
tokens identified as stopwords in a text sample.
It is defined as
\begin{equation}
    D_{\mathrm{stop}}(x_i) =
    \frac{C_{\mathrm{stop}}(x_i)}{N_i},
    \label{eq:stopword_density}
\end{equation}
where $N_i>0$. The resulting value ranges from
$0$ to $1$, with higher values indicating a
greater proportion of stopwords. Unlike the
absolute stopword count, stopword density
normalizes stopword usage by text length.

\paragraph{Matra Count.}
Matras are dependent vowel signs in Devanagari
script that represent vowel sounds associated
with consonants. Let
$x_i=(c_{i1},c_{i2},\ldots,c_{iL_i})$
denote the sequence of Unicode code points
in text sample $x_i$, where $L_i$ represents
the total number of code points. Let
$\mathcal{M}$ denote the predefined set of
Devanagari vowel-sign code points. The matra
count is defined as
\begin{equation}
    C_{\mathrm{matra}}(x_i) =
    \sum_{j=1}^{L_i}
    \mathbb{I}(c_{ij}\in\mathcal{M}),
    \label{eq:matra_count}
\end{equation}
where $\mathbb{I}(\cdot)$ is the indicator
function. This feature measures the frequency
of dependent vowel signs rather than
independent vowel letters. The same Unicode
character inventory is used consistently
across all text collections.

\paragraph{BERT-Based Semantic Distance.}
BERT-based semantic distance measures the
difference between the contextual representations
of a text sample and a reference text. Let
$h(x_i)\in\mathbb{R}^{d}$ denote the embedding
of text sample $x_i$ obtained using a specified
pretrained BERT-based encoder and pooling
strategy, and let $r_i$ denote its corresponding
reference text. Their cosine similarity
is defined as
\begin{equation}
    \operatorname{sim}(x_i,r_i) =
    \frac{
        h(x_i)^\top h(r_i)
    }{
        \|h(x_i)\|_2
        \|h(r_i)\|_2
    },
    \label{eq:bert_similarity}
\end{equation}
where both embeddings are assumed to be
nonzero. The corresponding cosine distance
is calculated as
\begin{equation}
    d_{\mathrm{BERT}}(x_i,r_i) =
    1-\operatorname{sim}(x_i,r_i).
    \label{eq:bert_distance}
\end{equation}
Lower distance values indicate greater
similarity between the contextual embeddings,
whereas higher values indicate greater
embedding dissimilarity. The encoder,
embedding extraction procedure, and
reference-text selection strategy must
be specified to ensure reproducibility.

\subsection{Quantifying Distributional Divergence Between Human-Written and LLM-Generated Texts}
\label{Appenix:divergence}
Since KL divergence is asymmetric, we compute both directions to provide complementary aspects of the distributional differences between human-written and LLM-generated texts. To quantify the distributional differences between the original text $s$
and its round-trip-translated counterpart $s^{rt}$, we compute the
Kullback-Leibler (KL) and Jensen-Shannon (JS) divergences over their
sentence representations. Let $\mathbf{h}_s$ and $\mathbf{h}_{s^{rt}}$
denote the sentence embeddings of the original and round-trip-translated
texts, respectively. We first transform these representations into
normalized probability distributions using the softmax function:
\begin{equation}
p = \operatorname{softmax}(\mathbf{h}_s),
\qquad
q = \operatorname{softmax}(\mathbf{h}_{s^{rt}}).
\end{equation}

The forward KL divergence quantifies the discrepancy between the
probability distributions induced by the original and round-trip-translated
sentence embeddings, with contributions weighted according to $p$.
Conversely, the reverse KL divergence measures the discrepancy in the
opposite direction, with contributions weighted according to $q$.
Since KL divergence is asymmetric, we consider both directions:
\begin{equation}
D_{\mathrm{KL}}(p\parallel q)
=
\sum_{i=1}^{d} p_i \log \frac{p_i}{q_i},
\end{equation}
\begin{equation}
D_{\mathrm{KL}}(q\parallel p)
=
\sum_{i=1}^{d} q_i \log \frac{q_i}{p_i},
\end{equation}
where $d$ denotes the dimensionality of the sentence embeddings.

We additionally compute the Jensen--Shannon (JS) divergence, which
provides a symmetric measure of the discrepancy between the two
probability distributions. Let $m = \frac{1}{2}(p+q)$ denote their
element-wise average. The JS divergence is defined as
\begin{equation}
D_{\mathrm{JS}}(p,q)
=
\frac{1}{2}D_{\mathrm{KL}}(p\parallel m)
+
\frac{1}{2}D_{\mathrm{KL}}(q\parallel m).
\end{equation}

These divergence measures characterize the differences between the
probability distributions derived from the original and
round-trip-translated sentence embeddings, providing complementary
indicators of representational changes introduced by the RTT process.

\begin{table}[h]
\centering
\small
\caption{Annotation scheme for human evaluation metrics, including score labels and descriptions.}
\label{tab:human_annotation}

\renewcommand{\arraystretch}{1}
\setlength{\tabcolsep}{4pt}
\rowcolors{2}{gray!30}{white}

\resizebox{\columnwidth}{!}{%
\begin{tabular}{l p{10.8cm}}
\toprule
\rowcolor{blue!50!black}
\textcolor{white}{\textbf{Metric}} &
\textcolor{white}{\textbf{Score Labels and Descriptions}} \\
\midrule

\textbf{Fluency} &
\textbf{3 -- Excellent:} No grammatical errors; text is clear and easy to understand. \newline
\textbf{2 -- Moderate:} Some grammatical errors or awkward phrasing, but the meaning remains understandable. \newline
\textbf{1 -- Poor:} Frequent grammatical errors; text is difficult to understand. \\
\midrule

\textbf{Origin} &
\textbf{1 -- Human-like:} Text appears natural and resembles human-written content. \newline
\textbf{0 -- LLM-like:} Text appears formulaic or resembles LLM-generated content. \\
\midrule

\textbf{Coherence} &
\textbf{1 -- Coherent:} Ideas are logically connected, consistent, and flow smoothly. \newline
\textbf{0 -- Incoherent:} Ideas are disconnected, inconsistent, or difficult to follow. \\
\midrule

\textbf{Factuality} &
\textbf{5 -- Completely accurate:} All relevant factual claims are correct and truthful. \newline
\textbf{4 -- Mostly accurate:} Main claims are correct, with minor omissions or inaccuracies. \newline
\textbf{3 -- Partially accurate:} Some claims are correct, but others are vague or imprecise. \newline
\textbf{2 -- Substantially inaccurate:} Noticeable factual errors or misleading statements are present. \newline
\textbf{1 -- Mostly incorrect:} Most claims are incorrect, irrelevant, or substantially misleading. \\

\bottomrule
\end{tabular}%
}
\end{table}

\newpage

\begin{table*}[t]
\centering
\setlength{\tabcolsep}{6pt}
\renewcommand{\arraystretch}{1.15}
\rowcolors{2}{gray!30}{white}

\caption{Configuration of the three TRIDENT classification variants, detailing the shared round-trip translation (RTT) and semantic similarity components, threshold-selection procedures, and classifier-specific hyperparameters. The static variant uses a validation-selected threshold, the dynamic variant estimates a decision boundary using a Gaussian Mixture Model (GMM), and the supervised variant employs logistic regression trained on labeled RTT-derived features.}
\label{tab:hyperparams}

\begin{tabular}{lll}
\toprule
\rowcolor{blue!50!black}
\textcolor{white}{\textbf{Variant}} &
\textcolor{white}{\textbf{Component}} &
\textcolor{white}{\textbf{Configuration}} \\
\midrule

\multicolumn{3}{l}{\textit{Shared Configuration}} \\
\midrule
All variants & RTT transformation & $L_1 \rightarrow L_2 \rightarrow L_1$ \\
All variants & Sentence encoder & XLM-RoBERTa (pretrained) \\
All variants & Similarity measure & Cosine similarity \\

\midrule
\multicolumn{3}{l}{\textit{Variant 1: TRIDENT-S (Static Threshold Classification)}} \\
\midrule
Static & Threshold selection & ROC-based validation \\
Static & Decision threshold & Dataset-specific validation threshold \\

\midrule
\multicolumn{3}{l}{\textit{Variant 2: TRIDENT-D (Dynamic GMM Threshold Classification)}} \\
\midrule
GMM & Number of components & 2 \\
GMM & Parameter estimation & Expectation--Maximization (EM) \\
GMM & Threshold selection & Midpoint of component means \\
GMM & Fitting data & Unlabeled development-set scores \\

\midrule
\multicolumn{3}{l}{\textit{Variant 3: TRIDENT-C (Supervised Logistic Regression)}} \\
\midrule
Logistic regression & Input features & Multi-distance RTT features \\
Logistic regression & Output function & Sigmoid \\
Logistic regression & Loss function & Binary cross-entropy \\
Logistic regression & Training data & Labeled training set \\
Logistic regression & Optimizer & AdamW \\
Logistic regression & Learning rate & $2 \times 10^{-5}$ \\
Logistic regression & Batch size & 8 \\
Logistic regression & Training epochs & 80 \\
Logistic regression & L2 regularization & $1 \times 10^{-4}$ \\

\bottomrule
\end{tabular}
\end{table*}

\begin{figure*}[t]
    \centering
\includegraphics[  width=\textwidth,
        height=\textheight,
        keepaspectratio]{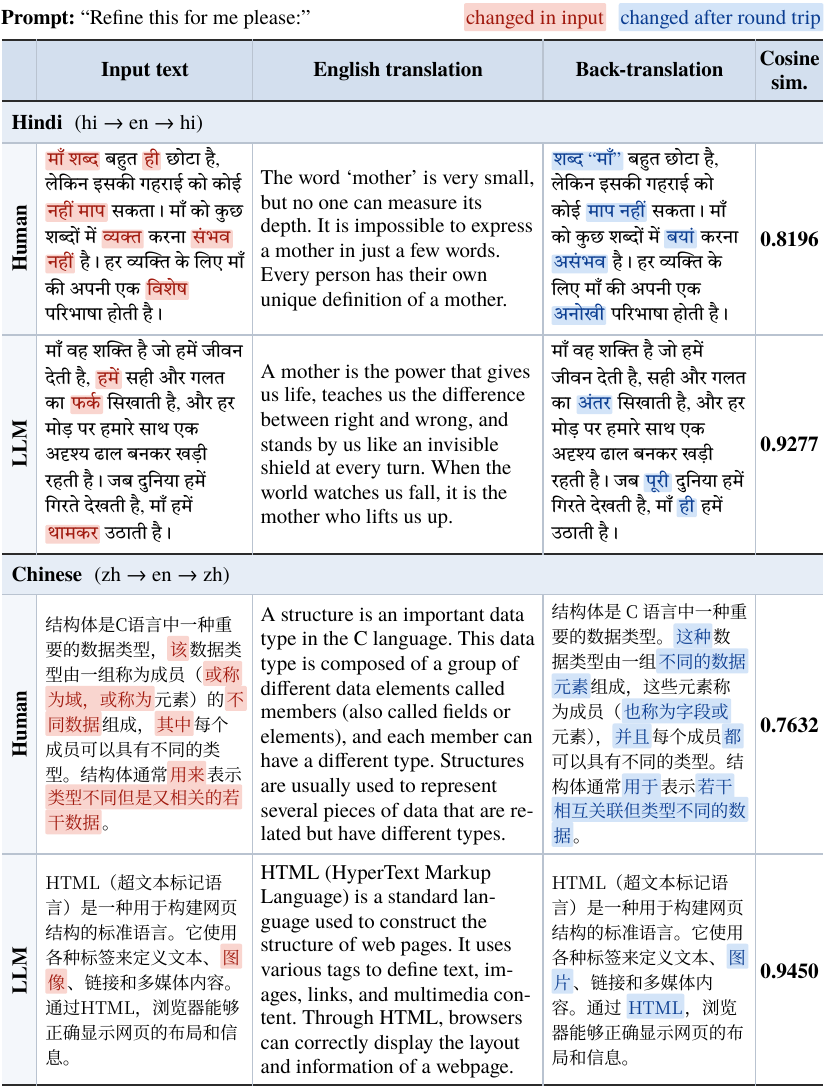}

\end{figure*}
\begin{figure*}[t]
    \centering
\includegraphics[
        width=\textwidth,
        height=0.85\textheight,
        keepaspectratio
    ]{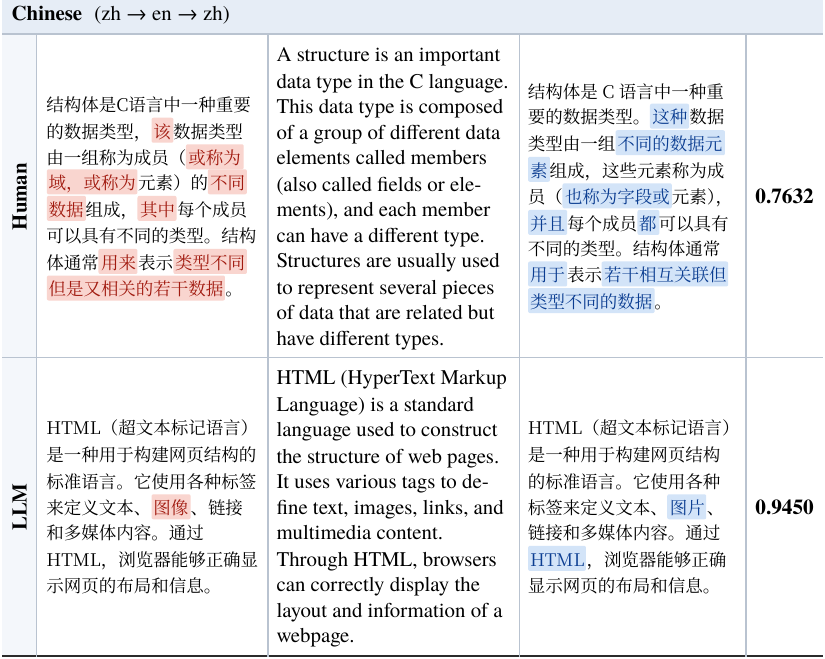}

    \caption{Example of a round-trip translation detection framework from 3 different languages. Proposed method, Deleted characters are marked in red, added characters are marked in blue, and unmodified
characters are in black. We exploit the difference in rewriting distance between human and AI texts for classification.
While the off-the-shelf GPT 3.5 model gives a different amount of rewrite for human and LLM texts (above), round-trip translation
from GPT 3.5 are much more separable (below). Close similarity using IF-IDF embedding.}
\label{proposed_method}
\end{figure*}

\begin{figure*}[t]
    \centering
\includegraphics[width=\textwidth,
        height=\textheight,
        keepaspectratio]{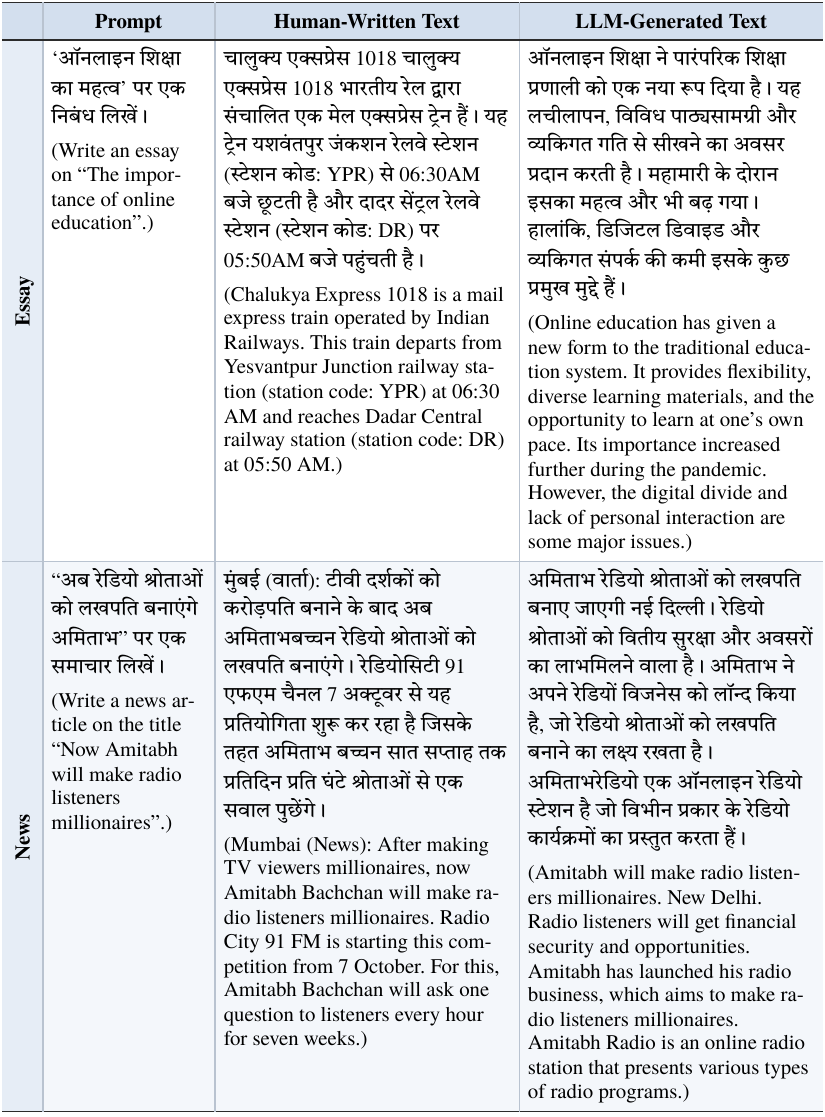}  
\end{figure*}

\begin{figure*}[t]
    \centering
\includegraphics[
        width=\textwidth,
        height=\textheight,
        keepaspectratio
    ]{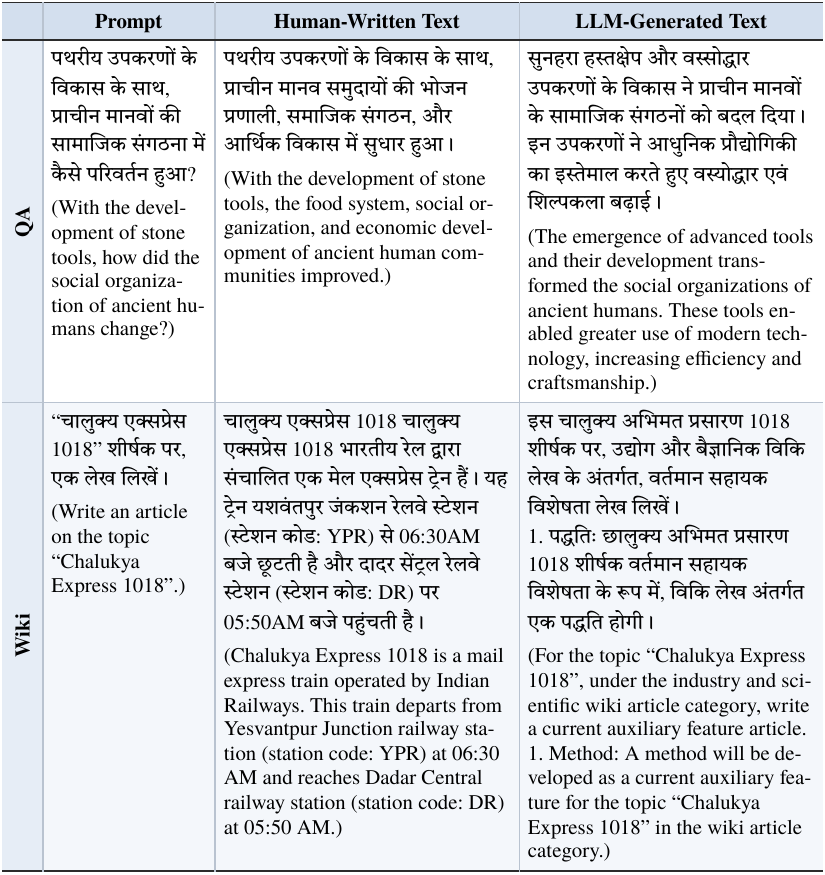}

    \caption{Representative example of a human-written Hindi passage and its corresponding LLM-generated counterpart in SHAKTI. The examples illustrate differences in content organisation, phrasing, and textual structure for the same topic.}
    \label{fig:dataset}
\end{figure*}

\begin{figure}[t]
\small
    \centering
    \includegraphics[width=0.9\linewidth]{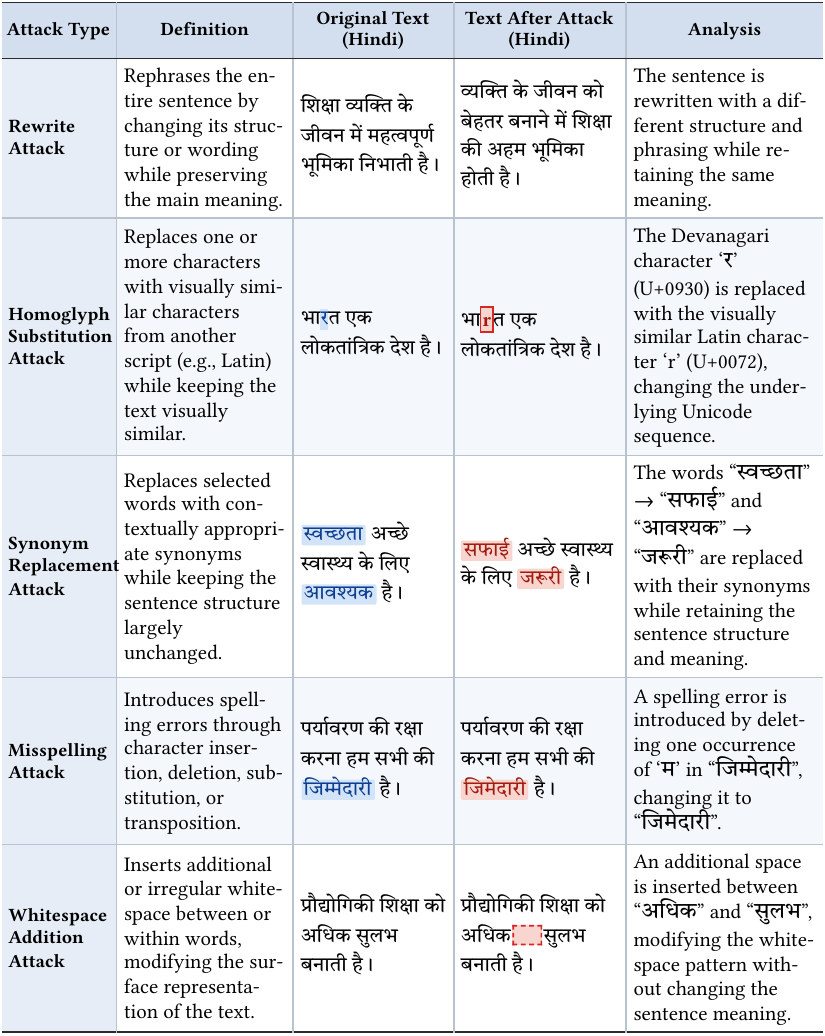} 
    \caption{Examples of five adversarial transformations applied to Hindi LLM-generated texts: rewriting, homoglyph substitution, synonym replacement, misspelling, and whitespace manipulation. For each transformation, the figure presents the original text, the perturbed text, and an explanation of the modification, illustrating how linguistic, character-level, and formatting changes can alter the textual representation while attempting to preserve meaning. }
    \label{fig:adversial_example}
\end{figure}

\end{document}